%% file: main.tex
\documentclass[10pt,twocolumn,letterpaper]{article}

\usepackage{cvpr}
\usepackage{times}
\usepackage{epsfig}
\usepackage{graphicx}
\usepackage{amsmath}
\usepackage{amssymb}
\usepackage{wrapfig}
\usepackage{diagbox}
\usepackage{verbatim}
\usepackage{xcolor}
\usepackage{floatrow}
\usepackage{booktabs}
\usepackage{multirow}
\usepackage[colorlinks,linkcolor=blue,bookmarks=false]{hyperref}
\usepackage{enumitem}
\usepackage{color}
\usepackage{placeins}
\usepackage[toc,page]{appendix}

\newcommand{\nickname}{RandLA-Net}

\definecolor{Gray}{gray}{0.85}
\newcolumntype{a}{>{\columncolor{Gray}}c}

\cvprfinalcopy 

\def\cvprPaperID{8374} 

\ifcvprfinal\fi
\begin{document}

\title{RandLA-Net: Efficient Semantic Segmentation of Large-Scale Point Clouds}

\author{Qingyong Hu\textsuperscript{1}, Bo Yang\textsuperscript{1\thanks{Corresponding author}*}, Linhai Xie\textsuperscript{1}, Stefano Rosa\textsuperscript{1}, Yulan Guo\textsuperscript{2,3}, \\Zhihua Wang\textsuperscript{1}, Niki Trigoni\textsuperscript{1}, Andrew Markham\textsuperscript{1} \\
\textsuperscript{1}University of Oxford, \textsuperscript{2}Sun  Yat-sen University, \textsuperscript{3}National University of Defense Technology\\
{\tt\small firstname.lastname@cs.ox.ac.uk}}

\maketitle

\begin{abstract}
    We study the problem of efficient semantic segmentation for large-scale 3D point clouds. By relying on expensive sampling techniques or computationally heavy pre/post-processing steps, most existing approaches are only able to be trained and operate over small-scale point clouds. In this paper, we introduce \textbf{\nickname{}}, an efficient and lightweight neural architecture to directly infer per-point semantics for large-scale point clouds. The key to our approach is to use random point sampling instead of more complex point selection approaches. Although remarkably computation and memory efficient, random sampling can discard key features by chance. To overcome this, we introduce a novel local feature aggregation module to progressively increase the receptive field for each 3D point, thereby effectively preserving geometric details. Extensive experiments show that our \nickname{} can process 1 million points in a single pass with up to 200$\times$ faster than existing approaches. Moreover, our \nickname{} clearly surpasses state-of-the-art approaches for semantic segmentation on two large-scale benchmarks Semantic3D and SemanticKITTI.  
\end{abstract}

\section{Introduction}
\label{sec:Intro}
\input{chapters/01_intro.tex}

\section{Related Work}
\input{chapters/02_Related_work.tex}

\section{\nickname{}}
\subsection{Overview}
\input{chapters/031_overview.tex}

\subsection{The quest for efficient sampling}
\label{Sub-sampling}
\input{chapters/032_sampling.tex}

\subsection{Local Feature Aggregation}
\label{LFA}
\input{chapters/033_LFA.tex}

\subsection{Implementation}
\label{subsec:Implementation}
\input{chapters/034_Implementation.tex}

\section{Experiments}
\input{chapters/040_experiments.tex}

\subsection{Efficiency of Random Sampling}
\label{sec:eff_sampling}
\input{chapters/041_Efficiency_Sampling.tex}

\subsection{Efficiency of \nickname{}}
\label{sec:eff_net}
\input{chapters/042_Efficiency_baseline.tex}

\subsection{Semantic Segmentation on Benchmarks}
\label{sec:sem_seg}
\input{chapters/043_Public_benchmark.tex}

\subsection{Ablation Study}
\label{sec:ablation}
\input{chapters/044_Ablation.tex}

\section{Conclusion}
\input{chapters/05_Conclusion.tex}

\clearpage
\noindent\textbf{Acknowledgments:} This work was partially supported by a China Scholarship Council (CSC) scholarship. Yulan Guo was supported by the National Natural Science Foundation of China (No. 61972435), Natural Science Foundation of Guangdong Province (2019A1515011271), and Shenzhen Technology and Innovation Committee.

{\small
\bibliographystyle{ieee_fullname}
\bibliography{egbib}
}

\clearpage
\input{chapters/06_supplementary.tex}

\end{document}

%% file: chapters/01_intro.tex
Efficient semantic segmentation of large-scale 3D point clouds is a fundamental and essential capability for real-time intelligent systems, such as autonomous driving and augmented reality. A key challenge is that the raw point clouds acquired by depth sensors are typically irregularly sampled, unstructured and unordered. Although deep convolutional networks show excellent performance in structured 2D computer vision tasks, they cannot be directly applied to this type of unstructured data.

\begin{figure}[t]
\label{fig:illustration}
\centering
\includegraphics[width=1.0\textwidth]{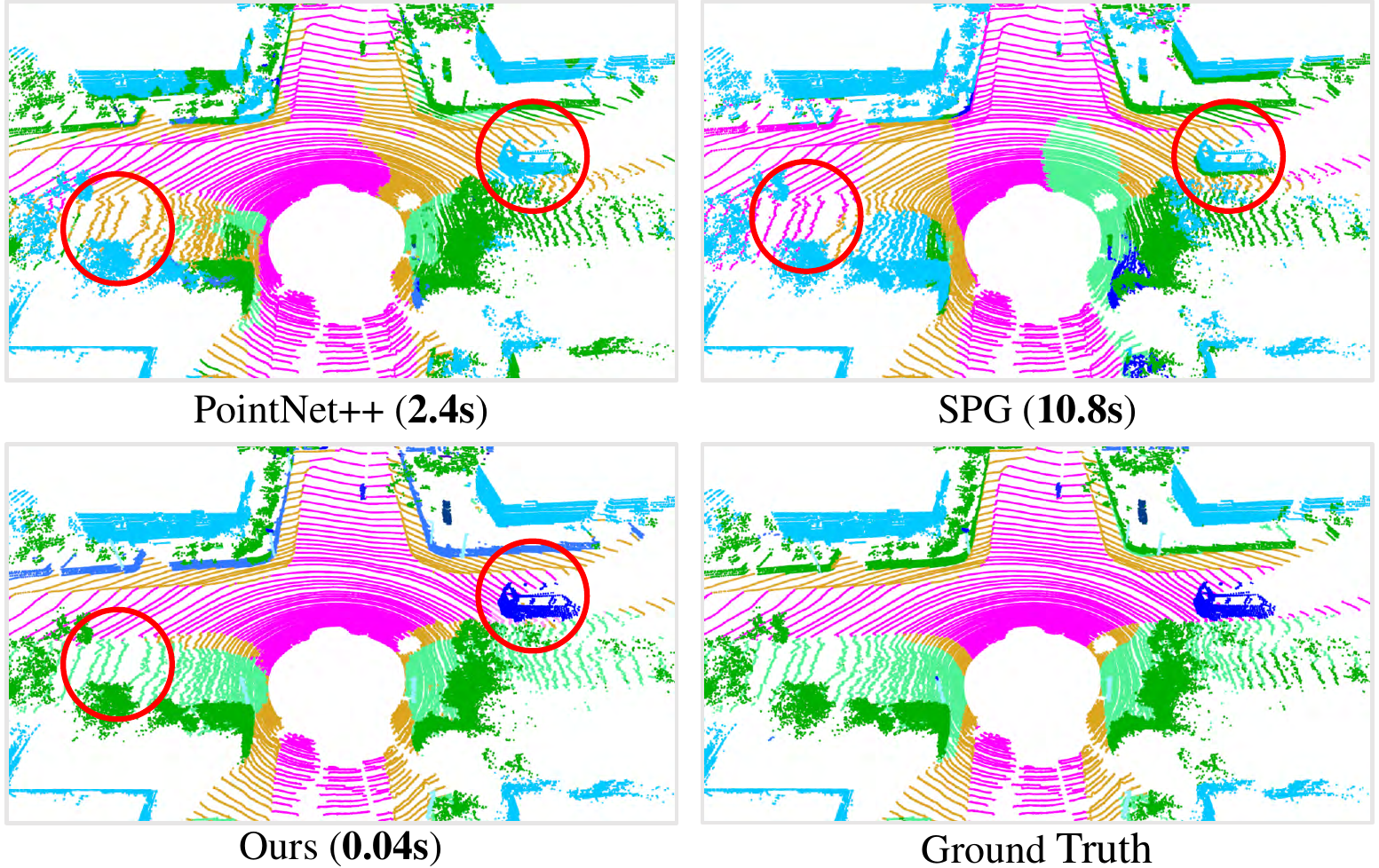}
\caption{Semantic segmentation results of PointNet++ \cite{qi2017pointnet++}, SPG \cite{landrieu2018large} and our approach on SemanticKITTI \cite{behley2019semantickitti}. Our \nickname{} takes only 0.04s to directly process a large point cloud with $10^5$ points over 150$\times$130$\times$10 meters in 3D space, which is up to 200$\times$ faster than SPG. Red circles highlight the superior segmentation accuracy of our approach.}
\vspace{-0.2cm}
\end{figure}

Recently, the pioneering work PointNet \cite{qi2017pointnet} has emerged as a promising approach for directly processing 3D point clouds. It learns per-point features using shared multilayer perceptrons (MLPs). This is computationally efficient but fails to capture wider context information for each point. To learn richer local structures, many dedicated neural modules have been subsequently and rapidly introduced. These modules can be generally categorized as: 1) neighbouring feature pooling \cite{qi2017pointnet++, so-net, RSNet, pointweb, zhang2019shellnet}, 2) graph message passing \cite{dgcnn, KCNet,local_spectral,GACNet, clusternet, HPEIN, Agglomeration}, 3) kernel-based convolution \cite{su2018splatnet, hua2018pointwise, wu2018pointconv, octree_guided, ACNN, Geo-CNN, thomas2019kpconv, mao2019interpolated}, and 4) attention-based aggregation \cite{xie2018attentional, PCAN, Yang2019ModelingPC, AttentionalPointNet}. Although these approaches achieve impressive results for object recognition and semantic segmentation, almost all of them are limited to extremely small 3D point clouds (e.g., 4k points or  1$\times$1 meter blocks) and cannot be directly extended to larger  point clouds (e.g., millions of points and up to 200$\times$200 meters) without preprocessing steps such as block partition. The reasons for this limitation are three-fold. 1) The commonly used point-sampling methods of these networks are either computationally expensive or memory inefficient. For example, the widely employed farthest-point sampling \cite{qi2017pointnet++} takes over 200 seconds to sample 10\% of 1 million points. 2) Most existing local feature learners usually rely on computationally expensive kernelisation or graph construction, thereby being unable to process massive number of points. 3) For a large-scale point cloud, which usually consists of hundreds of objects, the existing local feature learners are either incapable of capturing complex structures, or do so inefficiently, due to their limited size of receptive fields.

A handful of recent works have started to tackle the task of directly processing large-scale point clouds. SPG \cite{landrieu2018large} preprocesses the large point clouds as super graphs before applying neural networks to learn per super-point semantics. Both FCPN \cite{rethage2018fully} and PCT \cite{PCT} combine voxelization and point-level networks to process massive point clouds. Although they achieve decent segmentation accuracy, the preprocessing and voxelization steps are too computationally heavy to be deployed in real-time applications.

In this paper, we aim to design a memory and computationally efficient neural architecture, which is able to directly process large-scale 3D point clouds in a single pass, without requiring any pre/post-processing steps such as voxelization, block partitioning or graph construction. However, this task is extremely challenging as it requires: 1) a memory and computationally efficient sampling approach to progressively downsample large-scale point clouds to fit in the limits of current GPUs, and 2) an effective local feature learner to progressively increase the receptive field size to preserve complex geometric structures. To this end, we first systematically demonstrate that \textbf{random sampling} is a key enabler for deep neural networks to efficiently process large-scale point clouds. However, random sampling can discard key information, especially for objects with sparse points.
To counter the potentially detrimental impact of random sampling, we propose a new and efficient \textbf{local feature aggregation module} to capture complex local structures over progressively smaller point-sets.

Amongst existing sampling methods, farthest point sampling and inverse density sampling are the most frequently used for small-scale point clouds \cite{qi2017pointnet++, wu2018pointconv, li2018pointcnn, pointweb, Groh2018flexconv}. As point sampling is such a fundamental step within these networks, we investigate the relative merits of different approaches in Section \ref{Sub-sampling}, where we see that the commonly used sampling methods limit scaling towards large point clouds, and act as a significant bottleneck to real-time processing. However, we identify random sampling as by far the most suitable component for large-scale point cloud processing as it is fast and scales efficiently.  Random sampling is not without cost, because prominent point features may be dropped by chance and it cannot be used directly in existing networks without incurring a performance penalty.  To overcome this issue, we design a new local feature aggregation module in Section \ref{LFA}, which is capable of effectively learning complex local structures by progressively increasing the receptive field size in each neural layer. In particular, for each 3D point, we firstly introduce a local spatial encoding (LocSE) unit to explicitly preserve local geometric structures. Secondly, we leverage attentive pooling to automatically keep the useful local features. Thirdly, we stack multiple LocSE units and attentive poolings as a dilated residual block, greatly increasing the effective receptive field for each point. Note that all these neural components are implemented as shared MLPs, and are therefore remarkably memory and computational efficient.

Overall, being built on the principles of simple \textbf{rand}om sampling and an effective \textbf{l}ocal feature \textbf{a}ggregator, our efficient neural architecture, named \textbf{RandLA-Net}, not only is up to 200$\times$ faster than existing approaches on large-scale point clouds, but also surpasses the state-of-the-art semantic segmentation methods on both Semantic3D \cite{Semantic3D} and SemanticKITTI \cite{behley2019semantickitti} benchmarks. Figure \ref{fig:illustration} shows qualitative results of our approach.  Our key contributions are: 
\begin{itemize}[leftmargin=*]
    \item We analyse and compare existing sampling approaches, identifying random sampling as the most suitable component for efficient learning on large-scale point clouds.
    \item We propose an effective local feature aggregation module to preserve complex local structures by progressively increasing the receptive field for each point.
    \item We demonstrate significant memory and computational gains over baselines, and surpass the state-of-the-art semantic segmentation methods on multiple large-scale benchmarks.
\end{itemize}

%% file: chapters/02_Related_work.tex
To extract features from 3D point clouds, traditional approaches usually rely on hand-crafted features \cite{point_signatures, fast_hist, landrieu2017structured, hackel2016fast}. Recent learning based approaches \cite{guo2019deep, qi2017pointnet, Point_voxel_cnn} mainly include projection-based, voxel-based and point-based schemes which are outlined here.

\textbf{(1) Projection and Voxel Based Networks.}
To leverage the success of 2D CNNs, many works \cite{li2016vehicle_rss, chen2017multi, PIXOR, pointpillars} project/flatten 3D point clouds onto 2D images to address the task of object detection. However, geometric details may be lost during the projection. Alternatively, point clouds can be voxelized into 3D grids and then powerful 3D CNNs are applied in \cite{sparse, pointgrid, 4dMinkpwski, vvnet, Fast_point_rcnn}. Although they achieve leading results on semantic segmentation and object detection, their primary limitation is the heavy computation cost, especially when processing large-scale point clouds.

\textbf{(2) Point Based Networks.}
Inspired by PointNet/PointNet++ \cite{qi2017pointnet, qi2017pointnet++}, many recent works introduced sophisticated neural modules to learn per-point local features. These modules can be generally classified as 1) neighbouring feature pooling \cite{so-net, RSNet, pointweb, zhang2019shellnet}, 2) graph message passing \cite{dgcnn, KCNet,local_spectral,GACNet, clusternet, HPEIN, Agglomeration, Li_2019_ICCV}, 3) kernel-based convolution \cite{su2018splatnet, hua2018pointwise, wu2018pointconv, octree_guided, ACNN, Geo-CNN, thomas2019kpconv, mao2019interpolated}, and 4) attention-based aggregation \cite{xie2018attentional, PCAN, Yang2019ModelingPC, AttentionalPointNet}. Although these networks have shown promising results on small point clouds, most of them cannot directly scale up to large scenarios due to their high computational and memory costs. Compared with them, our proposed \nickname{} is distinguished in three ways: 1) it only relies on random sampling within the network, thereby requiring much less memory and computation; 2) the proposed local feature aggregator can obtain successively larger receptive fields by explicitly considering the local spatial relationship and point features, thus being more effective and robust for learning complex local patterns; 3) the entire network only consists of shared MLPs without relying on any expensive operations such as graph construction and kernelisation, therefore being superbly efficient for large-scale point clouds. 

\textbf{(3) Learning for Large-scale Point Clouds}.
SPG \cite{landrieu2018large} preprocesses the large point clouds as superpoint graphs to learn per super-point semantics. The recent FCPN \cite{rethage2018fully} and PCT \cite{PCT} apply both voxel-based and point-based networks to process the massive point clouds. However, both the graph partitioning and voxelisation are computationally expensive. In constrast, our \nickname{} is end-to-end trainable without requiring additional pre/post-processing steps.

%% file: chapters/031_overview.tex
\begin{figure}[htb]
\centering
\includegraphics[width=\textwidth]{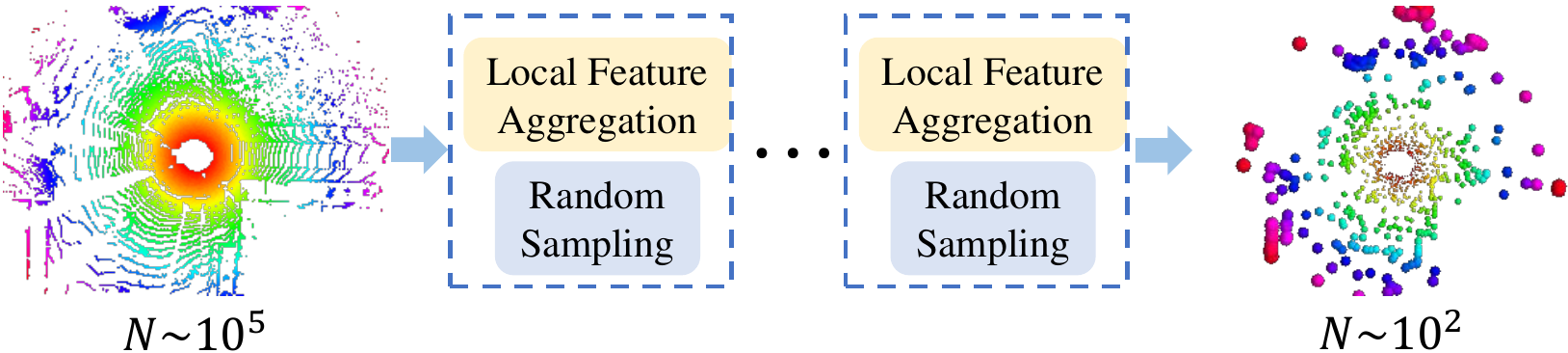}
\caption{In each layer of \nickname{}, the large-scale point cloud is significantly downsampled, yet is capable of retaining features necessary for accurate segmentation.}
\label{fig:sampling}
\end{figure}

As illustrated in Figure \ref{fig:sampling}, given a large-scale point cloud with millions of points spanning up to hundreds of meters, to process it with a deep neural network inevitably requires those points to be progressively and efficiently downsampled in each neural layer, without losing the useful point features. In our \nickname{}, we propose to use the simple and fast approach of random sampling to greatly decrease point density, whilst applying a carefully designed local feature aggregator to retain prominent features. This allows the entire network to achieve an excellent trade-off between efficiency and effectiveness.

\vspace{-0.2cm}

%% file: chapters/032_sampling.tex
\begin{figure*}[thb]
\centering
\includegraphics[width=\textwidth]{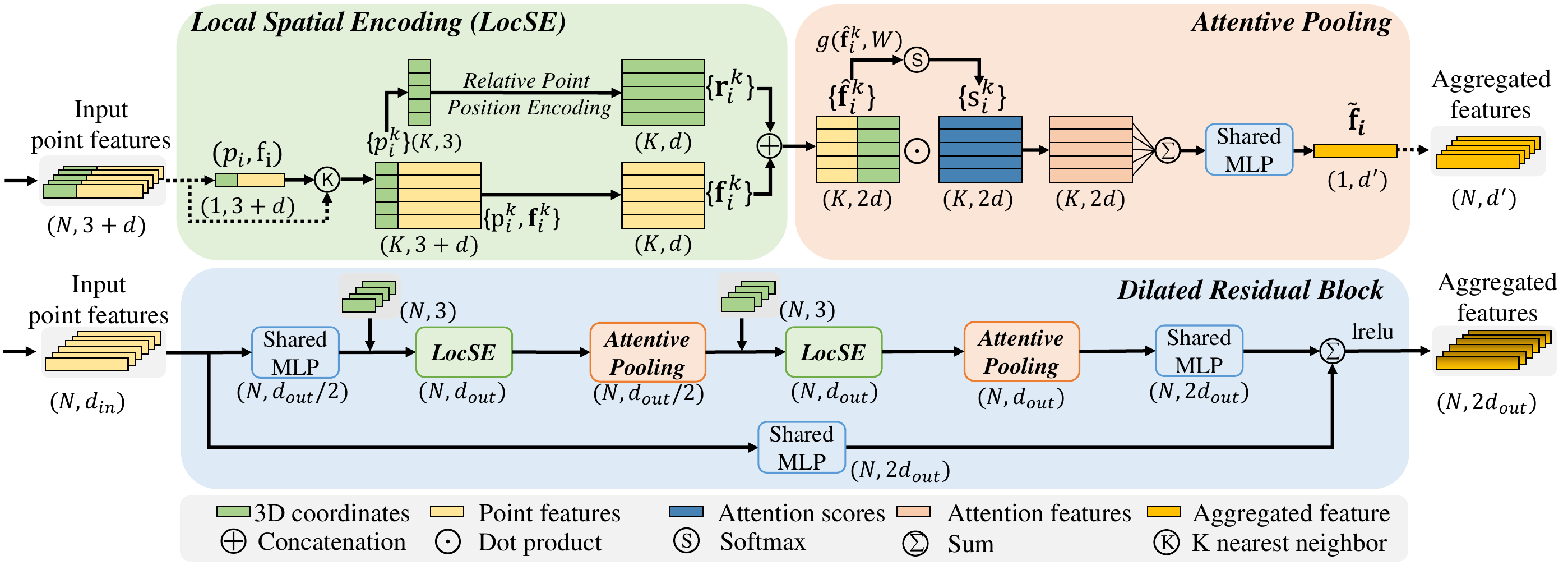}
\caption{The proposed local feature aggregation module. The top panel shows the location spatial encoding block that extracts features, and the attentive pooling mechanism that weights the most important neighbouring features, based on the local context and geometry. The bottom panel shows how two of these components are chained together, to increase the receptive field size, within a residual block.}
\label{fig:network}
\vspace{-0.2cm}
\end{figure*}

Existing point sampling approaches \cite{qi2017pointnet++, li2018pointcnn, Groh2018flexconv, learning2sample, concrete, wu2018pointconv} can be roughly classified into heuristic and learning-based approaches. However, there is still no standard sampling strategy that is suitable for large-scale point clouds. Therefore, we analyse and compare their relative merits and complexity as follows.\\ \vspace{-2mm}

\noindent\textbf{(1) Heuristic Sampling} 
\vspace{-2mm}
\begin{itemize}[leftmargin=*]
    \item\textit{Farthest Point Sampling (FPS):} In order to sample $K$ points from a large-scale point cloud $\boldsymbol{P}$ with $N$ points, FPS returns a reordering of the metric space $\{p_1 \cdots p_k \cdots p_K\}$, such that each $p_k$ is the farthest point from the first $k-1$ points. FPS is widely used in \cite{qi2017pointnet++, li2018pointcnn, wu2018pointconv} for semantic segmentation of small point sets. Although it has a good coverage of the entire point set, its computational complexity is $\mathcal{O}(N^2)$. For a large-scale point cloud ($N \sim 10^6$), FPS takes up to 200 seconds\footnote{We use the same hardware in Sec \ref{subsec:Implementation}, unless specified otherwise.}  to process on a single GPU. This shows that FPS is not suitable for large-scale point clouds.
    \item\textit{Inverse Density Importance Sampling (IDIS):} To sample $K$ points from $N$ points, IDIS reorders all $N$ points according to the density of each point, after which the top $K$ points are selected \cite{Groh2018flexconv}. Its computational complexity is approximately $\mathcal{O}(N)$. Empirically, it takes 10 seconds to process $10^6$ points. Compared with FPS, IDIS is more efficient, but also more sensitive to outliers. However, it is still too slow for use in a real-time system.
    \item \textit{Random Sampling (RS):} Random sampling uniformly selects $K$ points from the original $N$ points. Its computational complexity is $\mathcal{O}(1)$, which is agnostic to the total number of input points, i.e., it is constant-time and hence inherently scalable. Compared with FPS and IDIS, random sampling has the highest computational efficiency, regardless of the scale of input point clouds. It only takes 0.004s to process $10^6$ points.
\end{itemize}

\noindent\textbf{(2) Learning-based Sampling}
\vspace{-2.5mm}
\begin{itemize}[leftmargin=*]
\item \textit{Generator-based Sampling (GS):} GS \cite{learning2sample} learns to generate a small set of points to approximately represent the original large point set. However, FPS is usually used in order to match the generated subset with the original set at inference stage, incurring additional computation. In our experiments, it takes up to 1200 seconds to sample 10\% of $10^6$ points.
\item \textit{Continuous Relaxation based Sampling (CRS):}
CRS approaches \cite{concrete, Yang2019ModelingPC} use the reparameterization trick to relax the sampling operation to a continuous domain for end-to-end training.  In particular, each sampled point is learnt based on a weighted sum over the full point clouds. It results in a large weight matrix when sampling all the new points simultaneously with a one-pass matrix multiplication, leading to an unaffordable memory cost. For example, it is estimated to take more than a 300 GB memory footprint to sample 10\% of $10^6$ points.
\item \textit{Policy Gradient based Sampling (PGS):} PGS formulates the sampling operation as a Markov decision process \cite{show_attend}. It sequentially learns a probability distribution to sample the points. However, the learnt probability has high variance due to the extremely large exploration space when the point cloud is large. For example, to sample 10\% of $10^6$ points, the exploration space is $\mathrm{C}_{10^{6}}^{10^{5}}$ and it is unlikely to learn an effective sampling policy. We empirically find that the network is difficult to converge if PGS is used for large point clouds.
\end{itemize}

Overall, FPS, IDIS and GS are too computationally expensive to be applied for large-scale point clouds. CRS approaches have an excessive memory footprint and PGS is hard to learn. By contrast, random sampling has the following two advantages: 1) it is remarkably computational efficient as it is agnostic to the total number of input points, 2) it does not require extra memory for computation. Therefore, we safely conclude that random sampling is by far the most suitable approach to process large-scale point clouds compared with all existing alternatives. However, random sampling may result in many useful point features being dropped. To overcome it, we propose a powerful local feature aggregation module as presented in the next section.

%% file: chapters/033_LFA.tex
As shown in Figure \ref{fig:network}, our local feature aggregation module is applied to each 3D point in parallel and it consists of three neural units: 1) local spatial encoding (LocSE), 2) attentive pooling, and 3) dilated residual block.

\noindent\textbf{(1) Local Spatial Encoding}\\
Given a point cloud $\boldsymbol{P}$ together with per-point features (e.g., raw RGB, or intermediate learnt features), this local spatial encoding unit explicitly embeds the x-y-z coordinates of all neighbouring points, such that the corresponding point features are always aware of their relative spatial locations. This allows the LocSE unit to explicitly observe the local geometric patterns, thus eventually benefiting the entire network to effectively learn complex local structures. In particular, this unit includes the following steps:

\textit{Finding Neighbouring Points.} For the $i^{th}$ point, its neighbouring points are firstly gathered by the simple $K$-nearest neighbours (KNN) algorithm for efficiency. The KNN is based on the point-wise Euclidean distances.

\textit{Relative Point Position Encoding.} For each of the nearest $K$ points $\{p_i^1 \cdots p_i^k \cdots p_i^K\}$ of the center point $p_i$, we explicitly encode the relative point position as follows:
\begin{equation}
  \mathbf{r}_{i}^{k} = MLP\Big(p_i \oplus p_i^k \oplus (p_i-p_i^k) \oplus ||p_i-p_i^k||\Big)
\label{Eq1}
\end{equation}
where $p_i$ and $p_i^k$ are the x-y-z positions of points, $\oplus$ is the concatenation operation, and $||\cdot||$ calculates the Euclidean distance between the neighbouring and center points. It seems that $\mathbf{r}_{i}^{k}$ is encoded from redundant point positions. Interestingly, this tends to aid the network to learn local features and obtains good performance in practice.

\textit{Point Feature Augmentation.} For each neighbouring point $p_i^k$, the encoded relative point positions $\mathbf{r}_{i}^{k}$ are concatenated with its corresponding point features $\mathbf{f}_i^k$, obtaining an augmented feature vector $\mathbf{\hat{f}}_i^k$.

Eventually, the output of the LocSE unit is a new set of neighbouring features $\mathbf{\hat{F}}_i = \{\mathbf{\hat{f}}_i^1 \cdots \mathbf{\hat{f}}_i^k \cdots \mathbf{\hat{f}}_i^K \}$, which explicitly encodes the local geometric structures for the center point $p_i$. We notice that the recent work \cite{liu2019relation} also uses point positions to improve semantic segmentation. However, the positions are used to learn point scores in \cite{liu2019relation}, while our LocSE explicitly encodes the relative positions to augment the neighbouring point features.

\noindent\textbf{(2) Attentive Pooling}\\
This neural unit is used to aggregate the set of neighbouring point features $\mathbf{\hat{F}}_i$. Existing works \cite{qi2017pointnet++, li2018pointcnn} typically use max/mean pooling to hard integrate the neighbouring features, resulting in the majority of the information being lost. By contrast, we turn to the powerful attention mechanism to automatically learn important local features. In particular, inspired by \cite{Yang_ijcv2019}, our attentive pooling unit consists of the following steps.

\textit{Computing Attention Scores.} Given the set of local features $\mathbf{\hat{F}}_i = \{\mathbf{\hat{f}}_i^1 \cdots \mathbf{\hat{f}}_i^k \cdots \mathbf{\hat{f}}_i^K \}$, we design a shared function $g()$ to learn a unique attention score for each feature. Basically, the function $g()$ consists of a shared MLP followed by $softmax$. It is formally defined as follows:
\begin{equation}
  \mathbf{s}_{i}^{k} = g(\mathbf{\hat{f}}_i^k, \boldsymbol{W})
  \label{Eq2}
\end{equation}
where $\boldsymbol{W}$ is the learnable weights of a shared MLP.

\textit{Weighted Summation.} The learnt attention scores can be regarded as a soft mask which automatically selects the important features. Formally, these features are weighted summed as follows:
\begin{equation}
  \mathbf{\Tilde{f}}_{i} = \sum_{k=1}^{K}(\mathbf{\hat{f}}_i^k \cdot \mathbf{s}_{i}^{k})
\end{equation}

To summarize, given the input point cloud $\boldsymbol{P}$, for the $i^{th}$ point $p_i$, our LocSE and Attentive Pooling units learn to aggregate the geometric patterns and features of its $K$ nearest points, and finally generate an informative feature vector $\mathbf{\Tilde{f}}_{i}$.

\noindent\textbf{(3) Dilated Residual Block}\\
Since the large point clouds are going to be substantially downsampled, it is desirable to significantly increase the receptive field for each point, such that the geometric details of input point clouds are more likely to be reserved, even if some points are dropped. As shown in Figure \ref{fig:network}, inspired by the successful ResNet \cite{he2016deep} and the effective dilated networks \cite{DPC}, we stack multiple LocSE and Attentive Pooling units with a skip connection as a dilated residual block.

\begin{figure}[t]
\centering
\includegraphics[width=\textwidth]{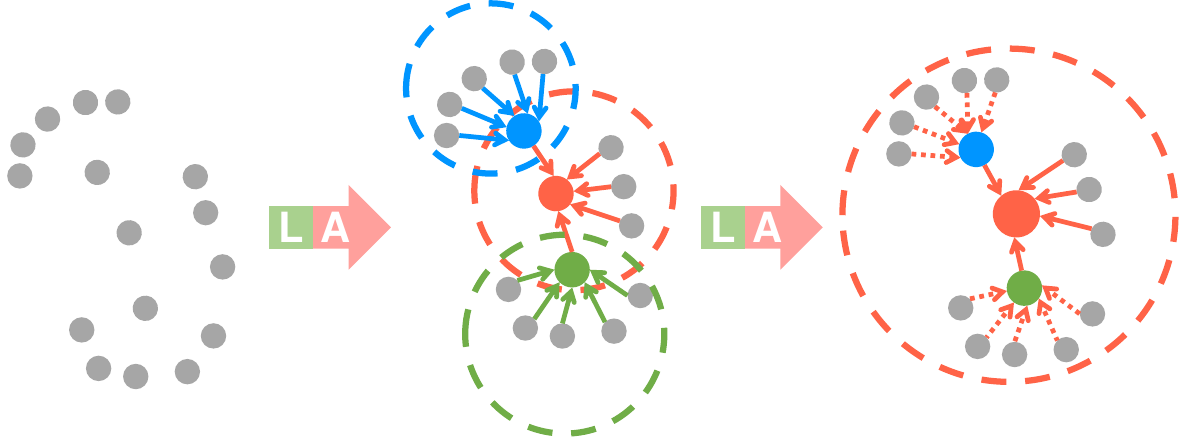}
\caption{Illustration of the dilated residual block which significantly increases the receptive field (dotted circle) of each point, colored points represent the aggregated features. L: Local spatial encoding, A: Attentive pooling.}
\label{fig:Residual}
\end{figure}

To further illustrate the capability of our dilated residual block, Figure \ref{fig:Residual} shows that the red 3D point observes $K$ neighbouring points after the first LocSE/Attentive Pooling operation, and then is able to receive information from up to $K^2$ neighbouring points i.e. its two-hop neighbourhood after the second. This is a cheap way of dilating the receptive field and expanding the effective neighbourhood through feature propagation. Theoretically, the more units we stack, the more powerful this block as its sphere of reach becomes greater and greater. However, more units would inevitably sacrifice the overall computation efficiency. In addition, the entire network is likely to be over-fitted. In our \nickname{}, we simply stack two sets of LocSE and Attentive Pooling as the standard residual block, achieving a satisfactory balance between efficiency and effectiveness. 

Overall, our local feature aggregation module is designed to effectively preserve complex local structures via explicitly considering neighbouring geometries and significantly increasing receptive fields. Moreover, this module only consists of feed-forward MLPs, thus being computationally efficient.

%% file: chapters/034_Implementation.tex
We implement \nickname{} by stacking multiple local feature aggregation modules and random sampling layers. The detailed architecture is presented in the Appendix. We use the Adam optimizer with default parameters. The initial learning rate is set as 0.01 and decreases by 5\% after each epoch. The number of nearest points $K$ is set as 16. To train our \nickname{} in parallel, we sample a fixed number of points ($\sim 10^5$) from each point cloud as the input. During testing, the whole raw point cloud is fed into our network to infer per-point semantics without pre/post-processing such as geometrical or block partition. All experiments are conducted on an NVIDIA RTX2080Ti GPU.

%% file: chapters/040_experiments.tex
%% file: chapters/041_Efficiency_Sampling.tex
In this section, we empirically evaluate the efficiency of existing sampling approaches including FPS, IDIS, RS, GS, CRS, and PGS, which have been discussed in Section \ref{Sub-sampling}. In particular, we conduct the following 4 groups of experiments.
\begin{itemize}[leftmargin=*]
    \item Group 1. Given a small-scale point cloud ($\sim 10^3$ points), we use each sampling approach to progressively downsample it. Specifically, the point cloud is downsampled by five steps with only 25\% points being retained in each step on a single GPU i.e. a four-fold decimation ratio. This means that there are only $\sim (1/4)^5 \times 10^3$ points left in the end. This downsampling strategy emulates the procedure used in PointNet++ \cite{qi2017pointnet++}. For each sampling approach, we sum up its time and memory consumption for comparison.
    \item Group 2/3/4. The total number of points are increased towards large-scale, i.e., around $10^4, 10^5$ and $10^6$ points respectively. We use the same five sampling steps as in Group 1. 
\end{itemize}

\textbf{Analysis.} Figure \ref{fig:sampling_comparison} compares the total time and memory consumption of each sampling approach to process different scales of point clouds. It can be seen that: 1) For small-scale point clouds ($\sim 10^3$), all sampling approaches tend to have similar time and memory consumption, and are unlikely to incur a heavy or limiting computation burden. 2) For large-scale point clouds ($\sim 10^6$), FPS/IDIS/GS/CRS/PGS are either extremely time-consuming or memory-costly. By contrast, random sampling has superior time and memory efficiency overall. This result clearly demonstrates that most existing networks \cite{qi2017pointnet++, li2018pointcnn, wu2018pointconv, liu2019relation, pointweb, Yang2019ModelingPC} are only able to be optimized on small blocks of point clouds primarily because they rely on the expensive sampling approaches. Motivated by this, we use the efficient random sampling strategy in our \nickname{}.

%% file: chapters/042_Efficiency_baseline.tex
\begin{figure}[t]
\centering
\includegraphics[width=\textwidth]{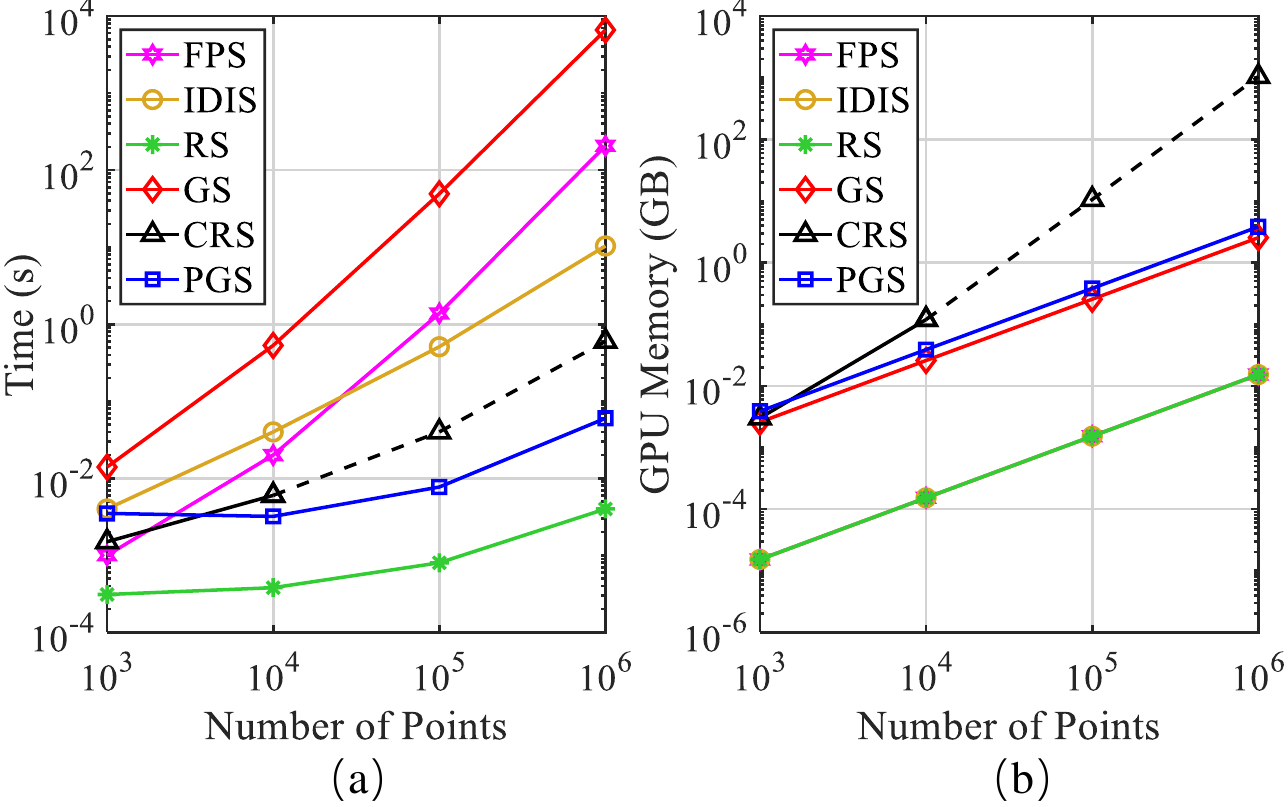}
\caption{Time and memory consumption of different sampling approaches. The dashed lines represent estimated values due to the limited GPU memory.}
\label{fig:sampling_comparison}
\vspace{-0.2cm}
\end{figure}

In this section, we systematically evaluate the overall efficiency of our \nickname{} on real-world large-scale point clouds for semantic segmentation. Particularly, we evaluate \nickname{} on the SemanticKITTI \cite{behley2019semantickitti} dataset, obtaining the total time consumption of our network on Sequence 08 which has 4071 scans of point clouds in total. We also evaluate the time consumption of recent representative works \cite{qi2017pointnet, qi2017pointnet++, li2018pointcnn, landrieu2018large, thomas2019kpconv} on the same dataset. For a fair comparison, we feed the same number of points (i.e., 81920) from each scan into each neural network.

In addition, we also evaluate the memory consumption of \nickname{} and the baselines. In particular, we not only report the total number of parameters of each network, but also measure the maximum number of 3D points each network can take as input in a single pass to infer per-point semantics. Note that, all experiments are conducted on the same machine with an AMD 3700X @3.6GHz CPU and an NVIDIA RTX2080Ti GPU.

\textbf{Analysis.} Table \ref{tab:efficiency} quantitatively shows the total time and memory consumption of different approaches. It can be seen that, 1) SPG \cite{landrieu2018large} has the lowest number of network parameters, but takes the longest time to process the point clouds due to the expensive geometrical partitioning and super-graph construction steps; 2) both PointNet++ \cite{qi2017pointnet++} and PointCNN \cite{li2018pointcnn} are also computationally expensive mainly because of the FPS sampling operation; 3) PointNet \cite{qi2017pointnet} and KPConv \cite{thomas2019kpconv} are unable to take extremely large-scale point clouds (e.g. $10^6$ points) in a single pass due to their memory inefficient operations. 4) Thanks to the simple random sampling together with the efficient MLP-based local feature aggregator, our \nickname{} takes the shortest time (185 seconds averaged by 4071 frames $\rightarrow$ roughly 22 FPS) to infer the semantic labels for each large-scale point cloud (up to $10^6$ points).

\begin{table}[tb]
\centering
\caption{The computation time, network parameters and maximum number of input points of different approaches for semantic segmentation on Sequence 08 of the SemanticKITTI \cite{behley2019semantickitti} dataset.}
\label{tab:efficiency}
\resizebox{\textwidth}{!}{%
\begin{tabular}{rccc}
\toprule[1.0pt]
 & \begin{tabular}[c]{@{}c@{}}Total time\\ (seconds)\end{tabular} & \begin{tabular}[c]{@{}c@{}}Parameters\\ (millions)\end{tabular} & \begin{tabular}[c]{@{}c@{}}Maximum inference\\ points (millions)\end{tabular} \\
\toprule[1.0pt]
PointNet (Vanilla) \cite{qi2017pointnet} & 192 & 0.8 & 0.49 \\
PointNet++ (SSG) \cite{qi2017pointnet++} & 9831 & 0.97 & 0.98 \\
PointCNN \cite{li2018pointcnn} & 8142 & 11 & 0.05 \\
SPG \cite{landrieu2018large} & 43584 & \textbf{0.25} & - \\
KPConv \cite{thomas2019kpconv} &717 &14.9 &0.54 \\
\textbf{\nickname{} (Ours)} & \textbf{185} & 1.24 & \textbf{1.03} \\
\toprule[1.0pt]
\end{tabular}%
}
\end{table}

%% file: chapters/043_Public_benchmark.tex
\begin{table*}[htb]
\centering
\caption{Quantitative results of different approaches on Semantic3D (reduced-8) \cite{Semantic3D}. Only the recent published approaches are compared. Accessed on 31 March 2020.}
\label{tab:reduced-8}
\resizebox{0.9\textwidth}{!}{%
\begin{tabular}{rcccccccccc}
\toprule[1.0pt]
 & mIoU (\%) & OA (\%) & man-made. & natural. & high veg. & low veg. & buildings & hard scape & scanning art. & cars \\
\toprule[1.0pt]
SnapNet\_ \cite{snapnet} & 59.1 & 88.6 & 82.0 & 77.3 & 79.7 & 22.9 & 91.1 & 18.4 & 37.3 & 64.4 \\
SEGCloud \cite{tchapmi2017segcloud} & 61.3 & 88.1 & 83.9 & 66.0 & 86.0 & 40.5 & 91.1 & 30.9 & 27.5 & 64.3 \\
RF\_MSSF \cite{RF_MSSF} & 62.7 & 90.3 & 87.6 & 80.3 & 81.8 & 36.4 & 92.2 & 24.1 & 42.6 & 56.6 \\
MSDeepVoxNet \cite{msdeepvoxnet} & 65.3 & 88.4 & 83.0 & 67.2 & 83.8 & 36.7 & 92.4 & 31.3 & 50.0 & 78.2 \\
ShellNet \cite{zhang2019shellnet} & 69.3 & 93.2 & 96.3 & 90.4 & 83.9 & 41.0 & 94.2 & 34.7 & 43.9 & 70.2 \\
GACNet \cite{GACNet} & 70.8 & 91.9 & 86.4 & 77.7 & \textbf{88.5} & \textbf{60.6} & 94.2 & 37.3 & 43.5 & 77.8 \\
SPG \cite{landrieu2018large} & 73.2 & 94.0 & \textbf{97.4} & \textbf{92.6} & 87.9 & 44.0 & 83.2 & 31.0 & 63.5 & 76.2 \\
KPConv \cite{thomas2019kpconv} & 74.6 & 92.9 & 90.9 & 82.2 & 84.2 & 47.9 & 94.9 & 40.0 & \textbf{77.3} & \textbf{79.7} \\
\textbf{\nickname{} (Ours)} & \textbf{77.4} & \textbf{94.8} & 95.6 & 91.4 & 86.6 & 51.5 & \textbf{95.7} & \textbf{51.5} & 69.8 & 76.8
\\
\bottomrule[1.0pt]
\end{tabular}%
}
\end{table*}

In this section, we evaluate the semantic segmentation of our \nickname{} on three large-scale public datasets: the outdoor Semantic3D \cite{Semantic3D} and SemanticKITTI \cite{behley2019semantickitti}, and the indoor S3DIS \cite{2D-3D-S}.

\begin{table*}[htb]
\centering
\caption{Quantitative results of different approaches on SemanticKITTI \cite{behley2019semantickitti}. Only the recent published methods are compared and all scores are obtained from the online single scan evaluation track. Accessed on 31 March 2020.}
\label{tab:SemanticKITTI}
\resizebox{\textwidth}{!}{%
\begin{tabular}{rcccccccccccccccccccccc}
\toprule[1.0pt]
Methods & Size & \rotatebox{90}{\textbf{mIoU(\%)}} & \rotatebox{90}{Params(M)}  & \rotatebox{90}{road} & \rotatebox{90}{sidewalk} & \rotatebox{90}{parking} & \rotatebox{90}{other-ground} & \rotatebox{90}{building} & \rotatebox{90}{car} & \rotatebox{90}{truck} & \rotatebox{90}{bicycle} & \rotatebox{90}{motorcycle} & \rotatebox{90}{other-vehicle} & \rotatebox{90}{vegetation} & \rotatebox{90}{trunk} & \rotatebox{90}{terrain} & \rotatebox{90}{person} & \rotatebox{90}{bicyclist} & \rotatebox{90}{motorcyclist} & \rotatebox{90}{fence} & \rotatebox{90}{pole} & \rotatebox{90}{traffic-sign} \\
\toprule[1.0pt]
PointNet \cite{qi2017pointnet} & \multirow{5}{*}{50K pts} & 14.6 & 3 & 61.6 & 35.7 & 15.8 & 1.4 & 41.4 & 46.3 & 0.1 & 1.3 & 0.3 & 0.8 & 31.0 & 4.6 & 17.6 & 0.2 & 0.2 & 0.0 & 12.9 & 2.4 & 3.7 \\
SPG \cite{landrieu2018large} &  & 17.4 & \textbf{0.25} & 45.0 & 28.5 & 0.6 & 0.6 & 64.3 & 49.3 & 0.1 & 0.2 & 0.2 & 0.8 & 48.9 & 27.2 & 24.6 & 0.3 & 2.7 & 0.1 & 20.8 & 15.9 & 0.8 \\
SPLATNet \cite{su2018splatnet} &  & 18.4 & 0.8 & 64.6 & 39.1 & 0.4 & 0.0 & 58.3 & 58.2 & 0.0 & 0.0 & 0.0 & 0.0 & 71.1 & 9.9 & 19.3 & 0.0 & 0.0 & 0.0 & 23.1 & 5.6 & 0.0 \\
PointNet++ \cite{qi2017pointnet++} &  & 20.1 & 6 & 72.0 & 41.8 & 18.7 & 5.6 & 62.3 & 53.7 & 0.9 & 1.9 & 0.2 & 0.2 & 46.5 & 13.8 & 30.0 & 0.9 & 1.0 & 0.0 & 16.9 & 6.0 & 8.9 \\
TangentConv \cite{tangentconv} &  & 40.9 & 0.4 & 83.9 & 63.9 & 33.4 & 15.4 & 83.4 & 90.8 & 15.2 & 2.7 & 16.5 & 12.1 & 79.5 & 49.3 & 58.1 & 23.0 & 28.4 & \textbf{8.1} & 49.0 & 35.8 & 28.5 \\
\toprule[1.0pt]
SqueezeSeg \cite{wu2018squeezeseg} & \multirow{5}{*}{\begin{tabular}[c]{@{}c@{}}64*2048\\ pixels\end{tabular}} & 29.5 & 1 & 85.4 & 54.3 & 26.9 & 4.5 & 57.4 & 68.8 & 3.3 & 16.0 & 4.1 & 3.6 & 60.0 & 24.3 & 53.7 & 12.9 & 13.1 & 0.9 & 29.0 & 17.5 & 24.5 \\
SqueezeSegV2 \cite{wu2019squeezesegv2} & & 39.7 & 1 & 88.6 & 67.6 & 45.8 & 17.7 & 73.7 & 81.8 & 13.4 & 18.5 & 17.9 & 14.0 & 71.8 & 35.8 & 60.2 & 20.1 & 25.1 & 3.9 & 41.1 & 20.2 & 36.3 \\
DarkNet21Seg \cite{behley2019semantickitti} & & 47.4 & 25 & 91.4 & 74.0 & 57.0 & 26.4 & 81.9 & 85.4 & 18.6 & \textbf{26.2} & 26.5 & 15.6 & 77.6 & 48.4 & 63.6 & 31.8 & 33.6 & 4.0 & 52.3 & 36.0 & 50.0 \\
DarkNet53Seg \cite{behley2019semantickitti} & & 49.9 & 50   & \textbf{91.8} & 74.6 & 64.8 & \textbf{27.9} & 84.1 & 86.4 & 25.5 & 24.5 & 32.7 & 22.6 & 78.3 & 50.1 & 64.0 & 36.2 & 33.6 & 4.7 & 55.0 & 38.9 & 52.2 \\
RangeNet53++ \cite{rangenet++} & & 52.2 & 50 & \textbf{91.8} & \textbf{75.2} & \textbf{65.0} & 27.8 & \textbf{87.4} & 91.4 & 25.7 & 25.7 & \textbf{34.4} & 23.0 & 80.5 & 55.1 & 64.6 &38.3  & 38.8  & 4.8 & \textbf{58.6} & 47.9 & \textbf{55.9} \\
\toprule[1.0pt]
\textbf{\nickname{} (Ours)} & 50K pts & \textbf{53.9} & 1.24 & 90.7 & 73.7 & 60.3 & 20.4 & 86.9 & \textbf{94.2} & \textbf{40.1} & 26.0 & 25.8 & \textbf{38.9} & \textbf{81.4} & \textbf{61.3} & \textbf{66.8} & \textbf{49.2} & \textbf{48.2} & 7.2 & 56.3 & \textbf{49.2} & 47.7 \\
\toprule[1.0pt]
\end{tabular}%
}
\end{table*}

\label{sec:result}
\begin{figure*}[hbt!]
\centering
\includegraphics[width=1\textwidth]{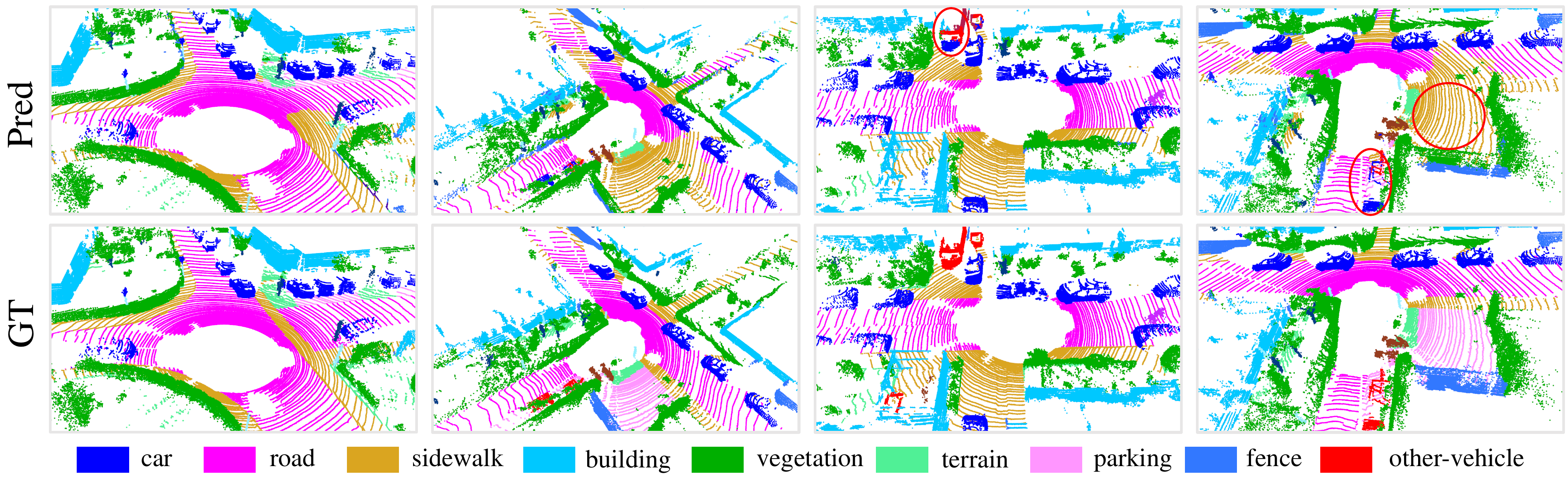}
\caption{Qualitative results of \nickname{} on the validation set of SemanticKITTI \cite{behley2019semantickitti}. Red circles show the failure cases.}
\label{fig:semanticKITTI}
\end{figure*}

\noindent\textbf{(1) Evaluation on Semantic3D.} The Semantic3D dataset \cite{Semantic3D} consists of 15 point clouds for training and 15 for online testing. Each point cloud has up to $10^{8}$ points, covering up to 160$\times$240$\times$30 meters in real-world 3D space. The raw 3D points belong to 8 classes and contain 3D coordinates, RGB information, and intensity. We only use the 3D coordinates and color information to train and test our \nickname{}. Mean Intersection-over-Union (mIoU) and Overall Accuracy (OA) of all classes are used as the standard metrics. For fair comparison, we only include the results of recently published strong baselines \cite{snapnet, tchapmi2017segcloud, RF_MSSF, msdeepvoxnet, zhang2019shellnet, GACNet, landrieu2018large} and the current state-of-the-art approach KPConv \cite{thomas2019kpconv}.

Table \ref{tab:reduced-8} presents the quantitative results of different approaches. \nickname{} clearly outperforms all existing methods in terms of both mIoU and OA. Notably, \nickname{} also achieves superior performance on six of the eight classes, except \textit{low vegetation} and \textit{scanning art.}.

\noindent\textbf{(2) Evaluation on SemanticKITTI.} SemanticKITTI \cite{behley2019semantickitti} consists of 43552 densely annotated LIDAR scans belonging to 21 sequences. Each scan is a large-scale point cloud with $\sim 10^{5}$ points and spanning up to 160$\times$160$\times$20 meters in 3D space. Officially, the sequences 00$\sim$07 and 09$\sim$10 (19130 scans) are used for training, the sequence 08 (4071 scans) for validation, and the sequences 11$\sim$21 (20351 scans) for online testing. The raw 3D points only have 3D coordinates without color information. The mIoU score over 19 categories is used as the standard metric.

Table 3 shows a quantitative comparison of our \nickname{} with two families of recent approaches, i.e. 1) point-based methods \cite{qi2017pointnet,landrieu2018large,su2018splatnet,qi2017pointnet++,tangentconv} and 2) projection based approaches \cite{wu2018squeezeseg, wu2019squeezesegv2,behley2019semantickitti, rangenet++}, and Figure \ref{fig:semanticKITTI} shows some qualitative results of \nickname{} on the validation split. It can be seen that our \nickname{} surpasses all point based approaches \cite{qi2017pointnet,landrieu2018large,su2018splatnet,qi2017pointnet++,tangentconv} by a large margin. We also 
outperform all projection based methods \cite{wu2018squeezeseg, wu2019squeezesegv2,behley2019semantickitti, rangenet++}, but not significantly, primarily because RangeNet++ \cite{rangenet++} achieves much better results on the small object category such as traffic-sign. However, our \nickname{} has $40\times$ fewer network parameters than RangeNet++ \cite{rangenet++} and is more computationally efficient as it does not require the costly steps of pre/post projection. 

\noindent\textbf{(3) Evaluation on S3DIS.} The S3DIS dataset \cite{2D-3D-S} consists of 271 rooms belonging to 6 large areas. Each point cloud is a medium-sized single room ($\sim$ 20$\times$15$\times$5 meters) with dense 3D points. To evaluate the semantic segmentation of our \nickname{}, we use the standard 6-fold cross-validation in our experiments. The mean IoU (mIoU), mean class Accuracy (mAcc) and Overall Accuracy (OA) of the total 13 classes are compared.

As shown in Table \ref{tab:s3dis}, our \nickname{} achieves on-par or better performance than state-of-the-art methods. Note that, most of these baselines \cite{qi2017pointnet++, li2018pointcnn, pointweb, zhang2019shellnet, dgcnn, chen2019lsanet} tend to use sophisticated but expensive operations or samplings to optimize the networks on small blocks (e.g., 1$\times$1 meter) of point clouds, and the relatively small rooms act in their favours to be divided into tiny blocks. By contrast, \nickname{} takes the entire rooms as input and is able to efficiently infer per-point semantics in a single pass.

\begin{table}[thb]
\centering
\caption{Quantitative results of different approaches on the S3DIS dataset \cite{2D-3D-S} (6-fold cross validation). Only the recent published methods are included.}
\label{tab:s3dis}
\resizebox{0.8\textwidth}{!}{%
\begin{tabular}{rccc}
\toprule[1.0pt]
 & OA(\%) & mAcc(\%) & mIoU(\%) \\
\toprule[1.0pt]
PointNet \cite{qi2017pointnet} & 78.6 & 66.2 & 47.6 \\
PointNet++ \cite{qi2017pointnet++} & 81.0 & 67.1 & 54.5 \\
DGCNN \cite{dgcnn} & 84.1 & - & 56.1 \\
3P-RNN \cite{3PRNN} & 86.9 & - & 56.3 \\
RSNet \cite{RSNet} & - & 66.5 & 56.5 \\
SPG \cite{landrieu2018large} & 85.5 & 73.0 & 62.1 \\
LSANet \cite{chen2019lsanet} & 86.8 & - & 62.2 \\
PointCNN \cite{li2018pointcnn} & 88.1 & 75.6 & 65.4 \\
PointWeb \cite{pointweb} & 87.3 & 76.2 & 66.7 \\
ShellNet \cite{zhang2019shellnet} & 87.1 & - & 66.8 \\
HEPIN \cite{HPEIN} & \textbf{88.2} & - & 67.8 \\
KPConv \cite{thomas2019kpconv} & - & 79.1 & \textbf{70.6} \\
\textbf{\nickname{} (Ours)} & 88.0 & \textbf{82.0} & 70.0 \\
\toprule[1.0pt]
\end{tabular}%
}
\end{table}
\vspace{-0.2cm}

%% file: chapters/044_Ablation.tex
Since the impact of random sampling is fully studied in Section \ref{sec:eff_sampling}, we  conduct the following ablation studies for our local feature aggregation module. All ablated networks are trained on sequences 00$\sim$07 and 09$\sim$10, and tested on the sequence 08 of SemanticKITTI dataset \cite{behley2019semantickitti}.

\textbf{(1) Removing local spatial encoding (LocSE).} This unit enables each 3D point to explicitly observe its local geometry. After removing locSE, we directly feed the local point features into the subsequent attentive pooling.

\textbf{(2$\sim$4) Replacing attentive pooling by max/mean/sum pooling.} The attentive pooling unit learns to automatically combine all local point features. By comparison, the widely used max/mean/sum poolings tend to hard select or combine features, therefore their performance may be sub-optimal.

\textbf{(5) Simplifying the dilated residual block.} The dilated residual block stacks multiple LocSE units and attentive poolings, substantially dilating the receptive field for each 3D point. By simplifying this block, we use only one LocSE unit and attentive pooling per layer, i.e. we do not chain multiple blocks as in our original \nickname{}.

Table \ref{tab:ablative} compares the mIoU scores of all ablated networks. From this, we can see that: 1) The greatest impact is caused by the removal of the chained spatial embedding and attentive pooling blocks. This is highlighted in Figure \ref{fig:Residual}, which shows how using two chained blocks allows information to be propagated from a wider neighbourhood, i.e. approximately $K^2$ points as opposed to  just $K$. This is especially critical with random sampling, which is not guaranteed to preserve a particular set of points. 2) The removal of the local spatial encoding unit shows the next greatest impact on performance, demonstrating that this module is necessary to effectively learn local and relative geometry context. 3) Removing the attention module diminishes performance by not being able to effectively retain useful features. From this ablation study, we can see how the proposed neural units complement each other to attain our state-of-the-art performance.

\begin{table}[htb]
\centering
\caption{The mean IoU scores of all ablated networks based on our full \nickname{}.}
\label{tab:ablative}
\resizebox{0.8\textwidth}{!}{%
\begin{tabular}{lc}
\toprule[1.0pt]
 & mIoU(\%) \\
\toprule[1.0pt]
(1) Remove local spatial encoding & 49.8 \\
(2) Replace with max-pooling & 55.2 \\
(3) Replace with mean-pooling & 53.4 \\
(4) Replace with sum-pooling & 54.3 \\
(5) Simplify dilated residual block & 48.8 \\
\textbf{(6) The Full framework (\nickname{})} & \textbf{57.1} \\
\toprule[1.0pt]
\end{tabular}%
}
\end{table}
\vspace{-0.2cm}

%% file: chapters/05_Conclusion.tex
In this paper, we demonstrated that it is possible to efficiently and effectively segment large-scale point clouds by using a lightweight network architecture. In contrast to most current approaches, that rely on expensive sampling strategies, we instead use random sampling in our framework to significantly reduce the memory footprint and computational cost. A local feature aggregation module is also introduced to effectively preserve useful features from a wide neighbourhood. Extensive experiments on multiple benchmarks demonstrate the high efficiency and the state-of-the-art performance of our approach. It would be interesting to extend our framework for the end-to-end 3D instance segmentation on large-scale point clouds by drawing on the recent work \cite{3dbonet} and also for the real-time dynamic point cloud processing \cite{liu2019meteornet}.

%% file: chapters/06_supplementary.tex
\begin{appendices}

\begin{figure*}[t]
\centering
\includegraphics[width=1\textwidth]{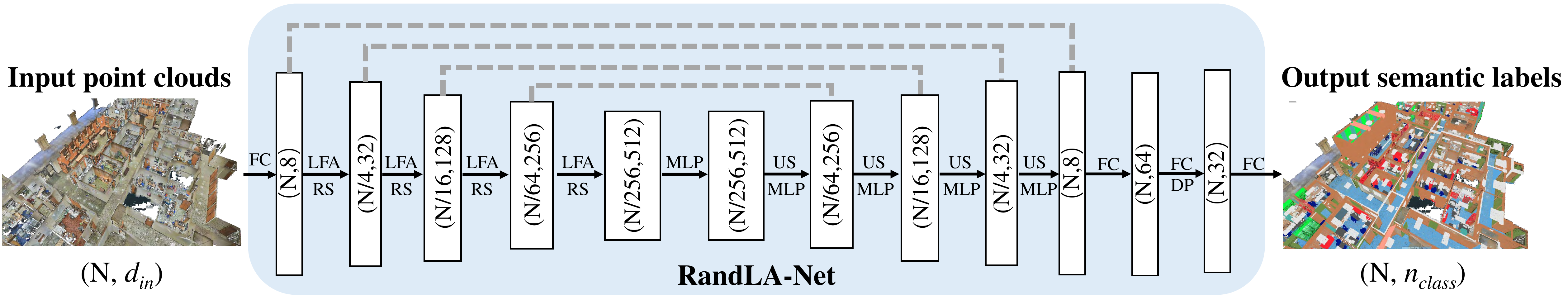}
\caption{The detailed architecture of our \nickname{}. $(N,D)$ represents the number of points and feature dimension respectively. FC: Fully Connected layer, LFA: Local Feature Aggregation, RS: Random Sampling, MLP: shared Multi-Layer Perceptron, US: Up-sampling, DP: Dropout.}
\label{fig:network-detailed}
\end{figure*}

\section{Details for the Evaluation of Sampling.}
We provide the implementation details of different sampling approaches evaluated in Section \ref{sec:eff_sampling}. To sample $K$ points (point features) from a large-scale point cloud $\boldsymbol{P}$ with $N$ points (point features):
\begin{enumerate}
\item \noindent\textit{Farthest Point Sampling (FPS):} We follow the implementation\footnote{\url{https://github.com/charlesq34/pointnet2}} provided by PointNet++ \cite{qi2017pointnet++}, which is also widely used in \cite{li2018pointcnn, wu2018pointconv, liu2019relation, chen2019lsanet, pointweb}. In particular, FPS is implemented as an operator running on GPU.
\item \noindent\textit{Inverse Density Importance Sampling (IDIS):} 
Given a point $p_i$, its density $\rho$ is approximated by calculating the summation of the distances between $p_i$ and its nearest $t$ points \cite{Groh2018flexconv}. Formally:
\begin{equation}
    \rho (p_{i})= \sum_{j=1}^{t} \left | \left | p_{i}-p_{i}^{j} \right | \right |, p_{i}^{j}\in \mathcal{N}(p_{i})
\end{equation}
\noindent where $p_{i}^{j}$ represents the coordinates (i.e. x-y-z) of the $j^{th}$ point of the neighbour points set $\mathcal{N}(p_{i})$, $t$ is set to 16. All the points are ranked according to the inverse density $\frac{1}{\rho}$ of points. Finally, the top $K$ points are selected.

\item \noindent\textit{Random Sampling (RS):} We implement random sampling with the python numpy package. Specifically, we first use the numpy function \textit{numpy.random.choice()} to generate $K$ indices. We then gather the corresponding spatial coordinates and per-point features from point clouds by using these indices.

\item \noindent\textit{Generator-based Sampling (GS):} The implementation follows the code\footnote{\url{https://github.com/orendv/learning_to_sample}} provided by \cite{learning2sample}. We first train a ProgressiveNet \cite{learning2sample} to transform the raw point clouds into ordered point sets according to their relevance to the task. After that, the first $K$ points are kept, while the rest is discarded.

\item \noindent\textit{Continuous Relaxation based Sampling (CRS):}  \textit{CRS} is implemented with the self-attended gumbel-softmax sampling \cite{concrete}\cite{Yang2019ModelingPC}. Given a point feature set $\boldsymbol{P} \in \mathbb{R}^{N\times (d+3)}$ with 3D coordinates and per point features, we firstly estimate a probability score vector $\mathbf{s} \in \mathbb{R}^{N}$ through a score function parameterized by a MLP layer, i.e., $\mathbf{s}=softmax(MLP(\boldsymbol{P}))$, which learns a categorical distribution. Then, with the Gumbel noise $\mathbf{g} \in \mathbb{R}^{N}$ drawn from the distribution $Gumbel(0, 1)$. Each sampled point feature vector $\mathbf{y} \in \mathbb{R}^{d+3}$ is calculated as follows:
\begin{equation}\label{eq: concrete sampling}
    \mathbf{y} = \sum_{i=1}^N 
    \dfrac{\exp{((log (\mathbf{s}^{(i)})+\mathbf{g}^{(i)})/\tau)} \boldsymbol{P}^{(i)}}
    {\sum_{j=1}^N \exp{((log (\mathbf{s}^{(j)}) + \mathbf{g}^{(j)})/\tau)}},
\end{equation}
where $\mathbf{s}^{(i)}$ and $\mathbf{g}^{(i)}$ indicate the $i^{th}$ element in the vector $\mathbf{s}$ and $\mathbf{g}$ respectively, $\boldsymbol{P}^{(i)}$ represents the $i^{th}$ row vector in the input matrix $\boldsymbol{P}$. $\tau > 0$ is the annealing temperature. When $\tau \rightarrow 0$, Equation~\ref{eq: concrete sampling} approaches the discrete distribution and samples each row vector in $\boldsymbol{P}$ with the probability $p(\mathbf{y}=\boldsymbol{P}^{(i)})=\mathbf{s}^{(i)}$.

\item \noindent\textit{Policy Gradients based Sampling (PGS):} Given a point feature set $\boldsymbol{P} \in \mathbb{R}^{N\times (d+3)}$ with 3D coordinates and per point features, we first predict a score $\mathbf{s}$ for each point, which is learnt by an MLP function, i.e., $\mathbf{s}=softmax(MLP(\boldsymbol{P}))+\mathbf{\epsilon}$, where $\mathbf{\epsilon} \in \mathbb{R}^N$ is a zero-mean Gaussian noise with the variance $\mathbf{\Sigma}$ for random exploration. After that, we sample $K$ vectors in $\boldsymbol{P}$ with the top $K$ scores. Sampling each point/vector can be regarded as an independent action and a sequence of them form a sequential Markov Decision Process (MDP) with the following policy function $\pi$:
\begin{equation}
    a_i\sim \pi(a|\boldsymbol{P}^{(i)}; \theta, \mathbf{s})
\end{equation}
where $a_i$ is the binary decision of whether to sample the $i^{th}$ vector in $\boldsymbol{P}$ and $\theta$ is the network parameter of the MLP.
Hence to properly improve the sampling policy with an underivable sampling process, we apply REINFORCE algorithm \cite{sutton2000policy} as the gradient estimator. The segmentation accuracy $R$ is applied as the reward value for the entire sampling process as $\mathcal{J}=R$. It is optimized with the following estimated gradients:
\begin{equation}\label{eq: reinforce}
  \begin{aligned}
    \dfrac{\partial \mathcal{J} }{\partial \theta} \approx  \dfrac{1}{M}\sum_{m=1}^M\sum_{i=1}^{N} \dfrac{\partial}{\partial \theta}\log \pi(a_i|\boldsymbol{P}^{(i)};\theta,\mathbf{\Sigma}) \times \\
    (R-b^c-b(\boldsymbol{P}^{(i)})),
    \end{aligned}
\end{equation}
where $M$ is the batch size, $b^c$ and $b(\boldsymbol{P}^{(i)})$ are two control variates \cite{mnih2014neural} for alleviating the high variance problem of policy gradients.
\end{enumerate}

\newpage
\section{Details of the Network Architecture}
\label{network_structure}
Figure \ref{fig:network-detailed} shows the detailed architecture of \nickname{}. The network follows the widely-used encoder-decoder architecture with skip connections. The input point cloud is first fed to a shared MLP layer to extract per-point features. Four encoding and decoding layers are then used to learn features for each point. At last, three fully-connected layers and a dropout layer are used to predict the semantic label of each point. The details of each part are as follows: \\

\noindent\textbf{Network Input:} The input is a large-scale point cloud with a size of $N\times d_{in}$ (the batch dimension is dropped for simplicity), where $N$ is the number of points, $d_{in}$ is the feature dimension of each input point. For both S3DIS \cite{2D-3D-S} and Semantic3D \cite{Semantic3D} datasets, each point is represented by its 3D coordinates and color information (i.e., x-y-z-R-G-B), while each point of the SemanticKITTI \cite{behley2019semantickitti} dataset is only represented by 3D coordinates.
\\

\noindent\textbf{Encoding Layers:} Four encoding layers are used in our network to progressively reduce the size of the point clouds and increase the per-point feature dimensions. Each encoding layer consists of a local feature aggregation module (Section \ref{LFA}) and a random sampling operation (Section \ref{Sub-sampling}). The point cloud is downsampled with a four-fold decimation ratio. In particular, only 25\% of the point features are retained after each layer, i.e.,  $(N\rightarrow \frac{N}{4}\rightarrow \frac{N}{16}\rightarrow \frac{N}{64}\rightarrow \frac{N}{256})$. Meanwhile, the per-point feature dimension is gradually increased each layer to preserve more information, i.e., $(8\rightarrow32\rightarrow 128\rightarrow 256\rightarrow 512)$. \\

\noindent\textbf{Decoding Layers:} Four decoding layers are used after the above encoding layers. For each layer in the decoder, we first use the KNN algorithm to find one nearest neighboring point for each query point, the point feature set is then upsampled through a nearest-neighbor interpolation. Next, the upsampled feature maps are concatenated with the intermediate feature maps produced by encoding layers through skip connections, after which a shared MLP is applied to the concatenated feature maps.\\

\noindent\textbf{Final Semantic Prediction:} The final semantic label of each point is obtained through three shared fully-connected layers ($N$, 64) $\rightarrow$ ($N$, 32) $\rightarrow$ ($N$, $n_{class}$) and a dropout layer. The dropout ratio is 0.5.\\

\noindent\textbf{Network Output:} The output of \nickname{} is the predicted semantics of all points, with a size of $ N\times n_{class}$, where $n_{class}$ is the number of classes.

\section{Additional Ablation Studies on LocSE}
In Section \ref{LFA}, we encode the relative point position based on the following equation:
\begin{equation}
  \mathbf{r}_{i}^{k} = MLP\Big(p_i \oplus p_i^k \oplus (p_i-p_i^k) \oplus ||p_i-p_i^k||\Big)
\label{Eq1_sup}
\end{equation}

We further investigate the effects of different spatial information in our framework. Particularly, we conduct the following more ablative experiments for LocSE:
\begin{itemize}
    \item 1) Encoding the coordinates of the point $p_i$ only.
    \item 2) Encoding the coordinates of  neighboring points $p_i^k$ only. 
    \item 3) Encoding the coordinates of the point $p_i$ and its neighboring points $p_i^k$. 
    \item 4) Encoding the coordinates of the point $p_i$, the neighboring points $p_i^k$, and Euclidean distance $||p_i-p_i^k||$.
    \item 5) Encoding the coordinates of the point $p_i$, the neighboring points $p_i^k$, and the relative position $p_{i}-p_{i}^{k}$.
\end{itemize}

\begin{table}[thb]
\centering
\caption{The mIoU result of \nickname{} by encoding different kinds of spatial information.}
\label{tab:LocSE_ablation}
\resizebox{0.9\textwidth}{!}{%
\begin{tabular}{lc}
\toprule[1.0pt]
LocSE & mIoU(\%) \\
\toprule[1.0pt]
(1) $(p_{i})$ &  45.5\\
(2) $(p_{i}^{k})$ &  47.7\\
(3) $(p_{i}, p_{i}^{k})$ & 49.1 \\
(4) $(p_{i}, p_{i}^{k}, ||p_i-p_i^k||)$ & 50.5 \\
(5) $(p_{i}, p_{i}^{k}, p_{i}-p_{i}^{k})$ &  53.6\\
(6) $(p_{i}, p_{i}^{k}, p_{i}-p_{i}^{k}, ||p_i-p_i^k||)$ (\textbf{The Full Unit}) & 54.3 \\ 
\toprule[1.0pt]
\end{tabular}
}
\end{table}

Table \ref{tab:LocSE_ablation} compares the mIoU scores of all ablated networks on the SemanticKITTI \cite{behley2019semantickitti} dataset. We can see that: 1) Explicitly encoding all spatial information leads to the best mIoU performance. 2) The relative position $p_{i}-p_{i}^{k}$ plays an important role in this component, primarily because the relative point position enables the network to be aware of the local geometric patterns. 3) Only encoding the point position $p_i$ or $p_i^k$ is unlikely to improve the performance, because the relative local geometric patterns are not explicitly encoded. \\

\section{Additional Ablation Studies on Dilated Residual Block} 
In our \nickname{}, we stack two LocSE and Attentive Pooling units as the standard dilated residual block to gradually increase the receptive field. To further evaluate how the number of aggregation units in the dilated residual block impact the entire network, we conduct the following two more groups of experiments.
\begin{itemize}
    \item 1) We simplify the dilated residual block by using only one LocSE unit and attentive pooling.
    \item 2) We add one more LocSE unit and attentive pooling, i.e., there are three aggregation units chained together.
\end{itemize}

\begin{table}[H]
\centering
\caption{The mIoU scores of \nickname{} regarding different number of aggregation units in a residual block.}
\label{tab:num_aggregation}
\resizebox{0.8\textwidth}{!}{%
\begin{tabular}{lc}
\toprule[1.0pt]
Dilated residual block & mIoU(\%) \\
\toprule[1.0pt]
(1) one aggregation unit &  49.8 \\ 
(2) three aggregation units &  51.1 \\ 
(3) two aggregation units (\textbf{The Standard Block} ) & 54.3 \\ 
\toprule[1.0pt]
\end{tabular}%
}
\end{table}

Table \ref{tab:num_aggregation} shows the mIoU scores of different ablated networks on the validation split of the SemanticKITTI \cite{behley2019semantickitti} dataset. It can be seen that: 1) Only one aggregation unit in the dilated residual block leads to a significant drop in segmentation performance, due to the limited receptive field. 2) Three aggregation units in each block do not improve the accuracy as expected. This is because the significantly increased receptive fields and the large number of trainable parameters tend to be overfitted.

\section{Visualization of Attention Scores}
\begin{figure}[thb]
\centering
\includegraphics[width=1\textwidth]{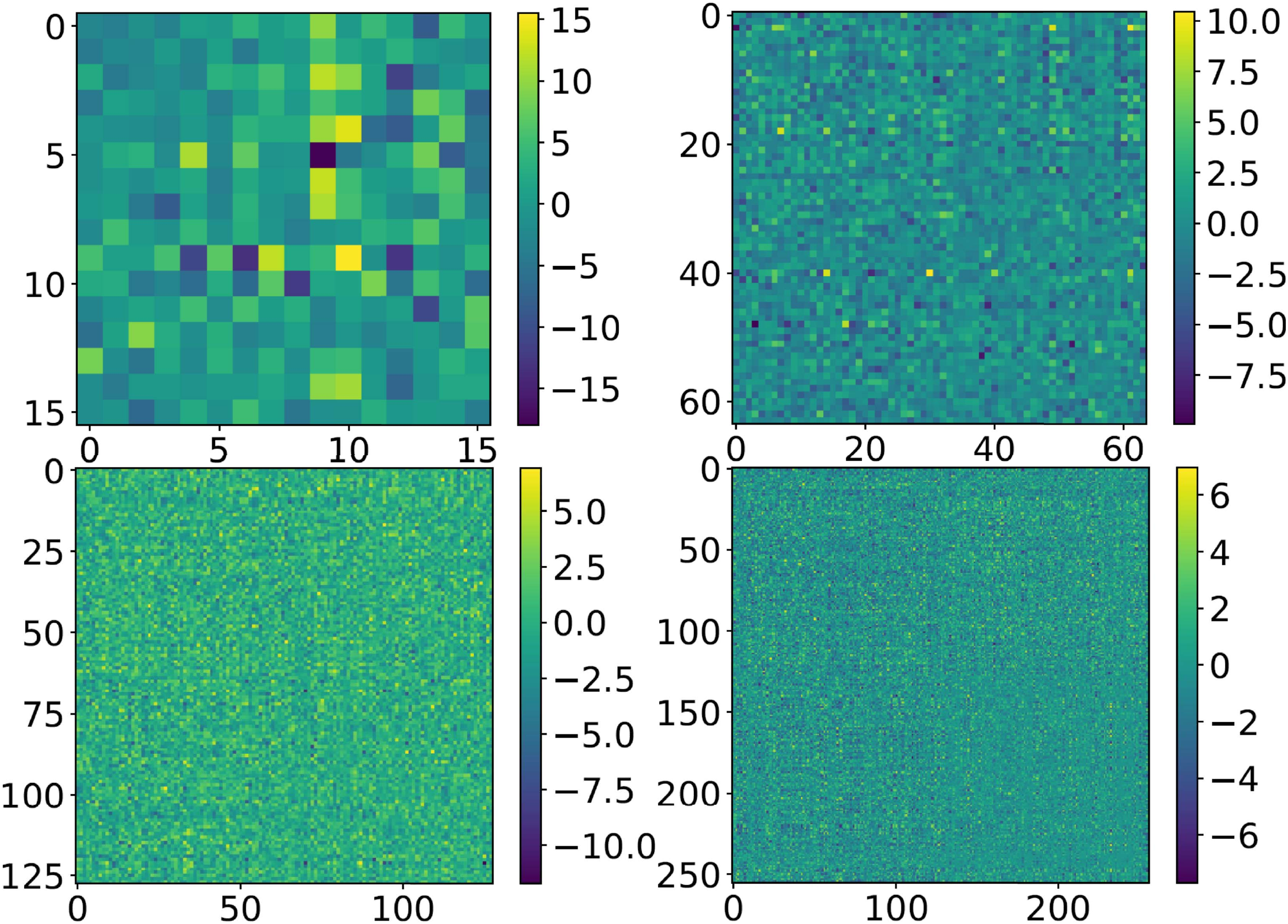}
\caption{Visualization of the learned attention matrix in different layers. From top left to bottom right: 16$\times$16 attention matrix, 64$\times$64 attention matrix, 128$\times$128 attention matrix, 256$\times$256 attention matrix. The yellow color represents higher attention scores.}
\label{attention_matrix}
\end{figure}

To better understand the attentive pooling, it is desirable to visualize the learned attention scores. However, since the attentive pooling operates on a relatively small local point set (i.e., $K$=16), it is hardly able to recognize meaningful shapes from such small local regions. Alternatively, we visualize the learned attention weight matrix $W$ defined in Equation \ref{Eq2} in each layer. As shown in Figure \ref{attention_matrix}, the attention weights have large values in the first encoding layers, then gradually become smooth and stable in subsequent layers. This shows that the attentive pooling tends to choose prominent or key point features at the beginning. After the point cloud being significantly downsampled, the attentive pooling layer tends to retain the majority of those point features.
\vspace{-0.2cm}

\section{Additional Results on Semantic3D}
\label{Add_semantic3d}
\begin{figure*}[t]
\centering
\includegraphics[width=0.47\textwidth]{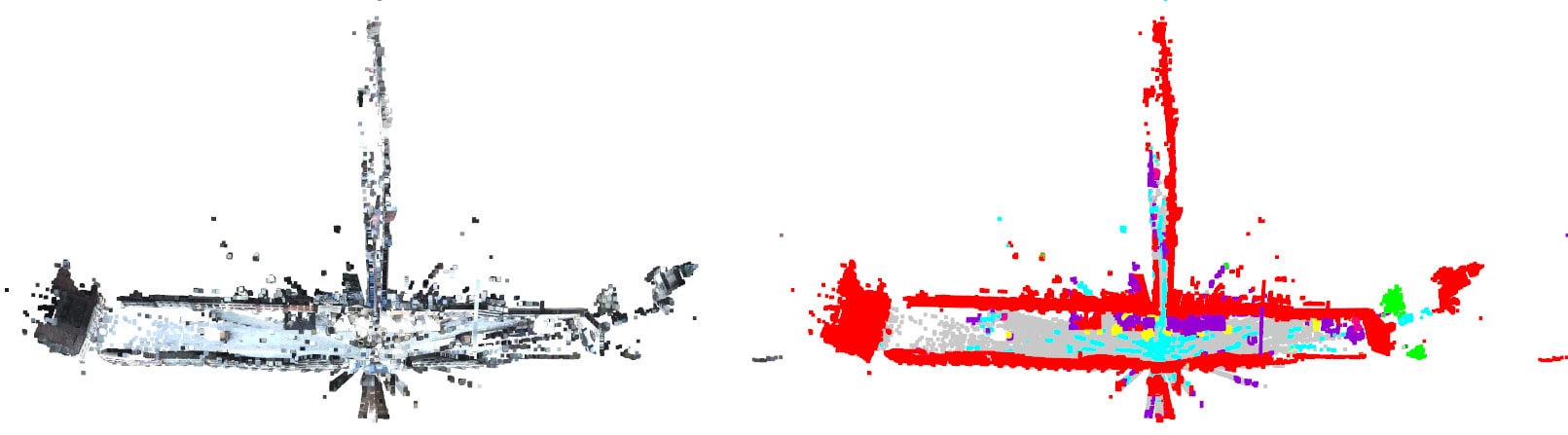}
\includegraphics[width=0.47\textwidth]{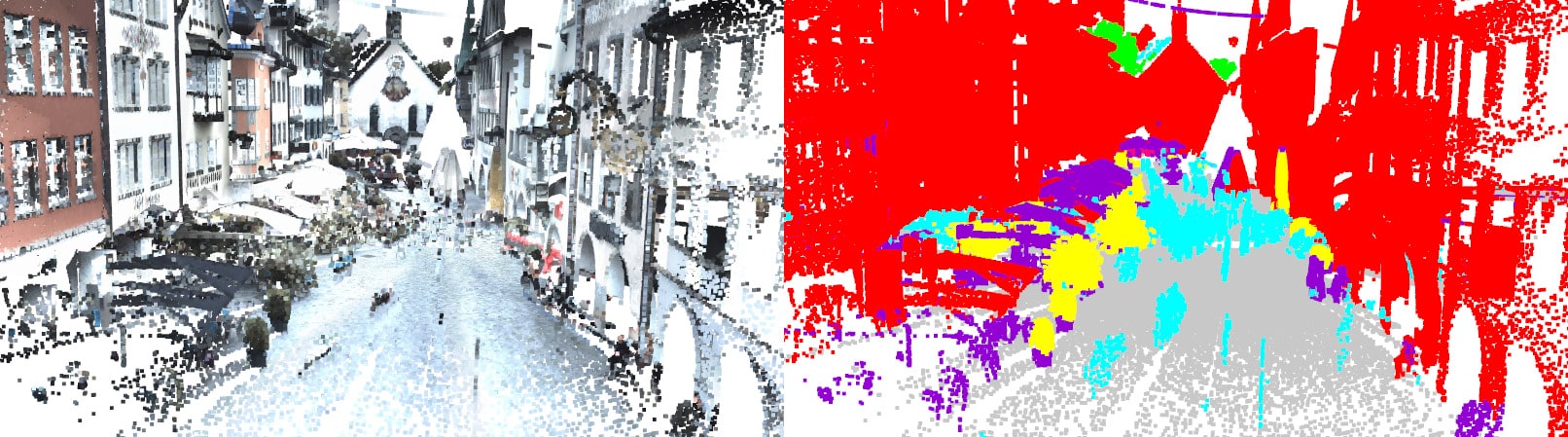}
\vspace{4pt}
\vspace{4pt}
\includegraphics[width=0.47\textwidth]{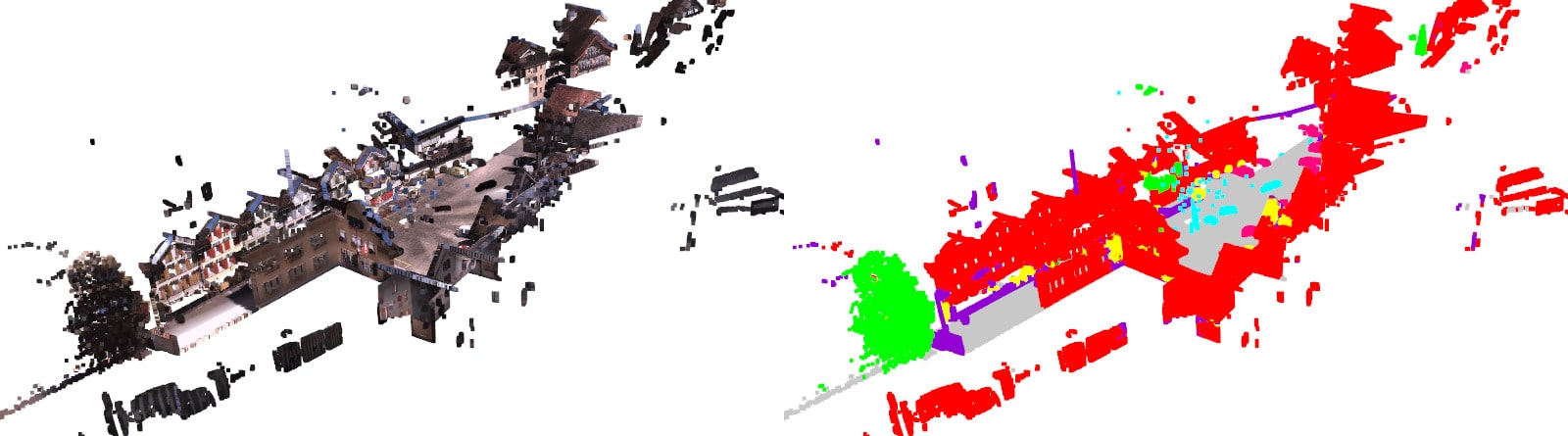}
\includegraphics[width=0.47\textwidth]{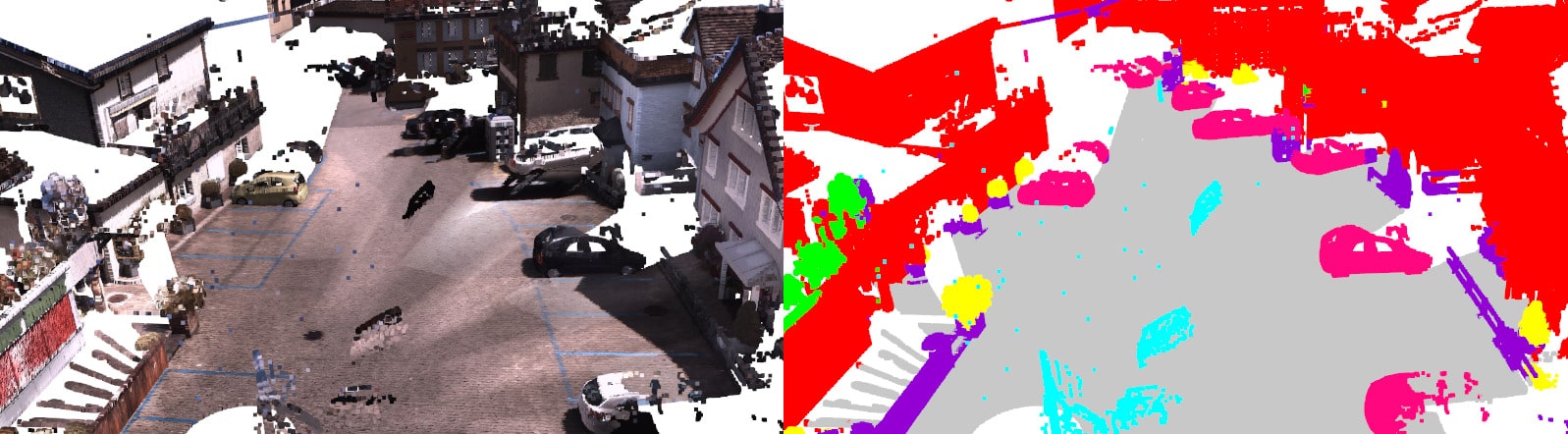}
\vspace{4pt}
\vspace{4pt}
\includegraphics[width=0.47\textwidth]{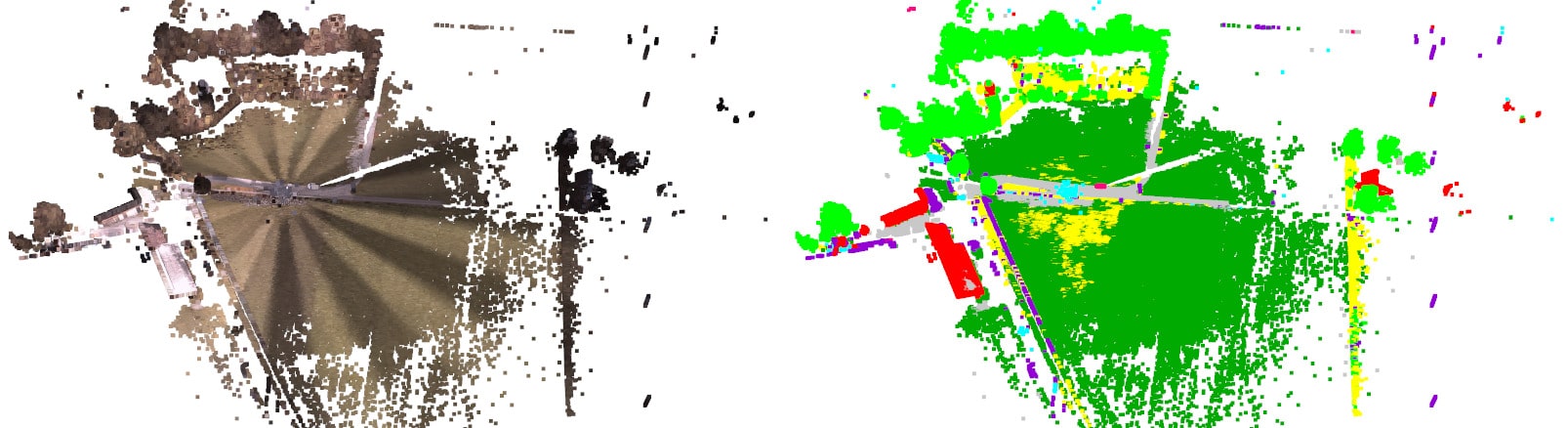}
\includegraphics[width=0.47\textwidth]{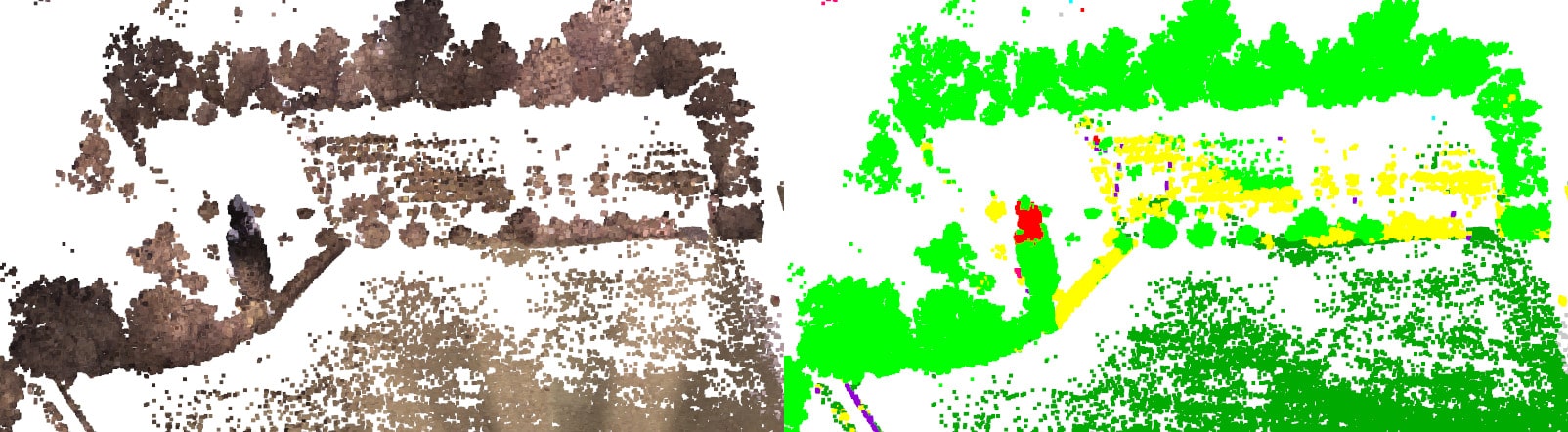}
\vspace{4pt}
\vspace{4pt}
\includegraphics[width=0.47\textwidth]{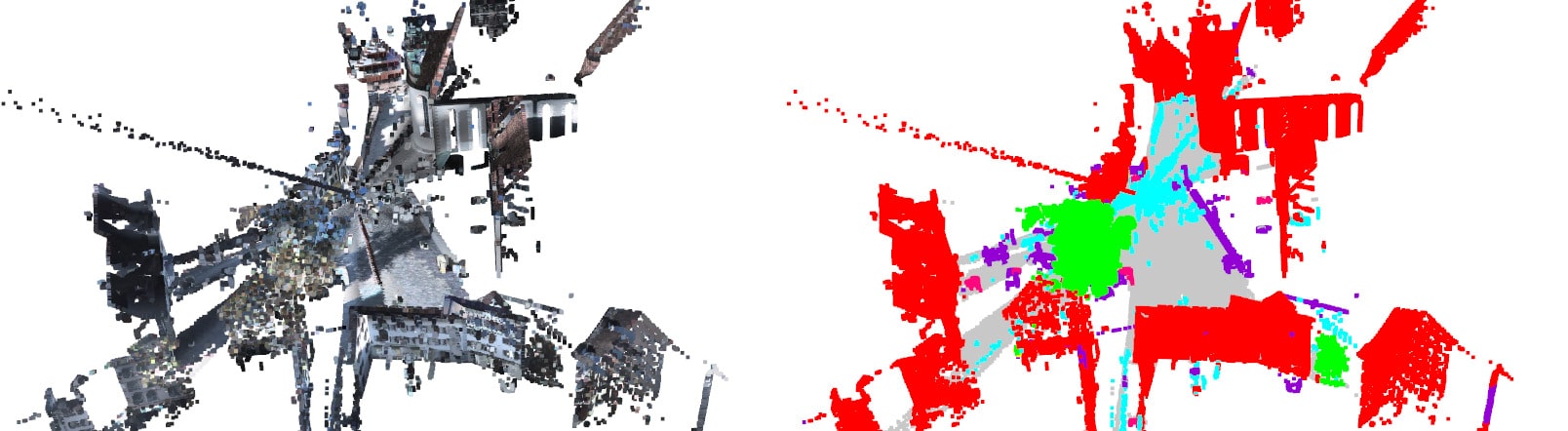}
\includegraphics[width=0.47\textwidth]{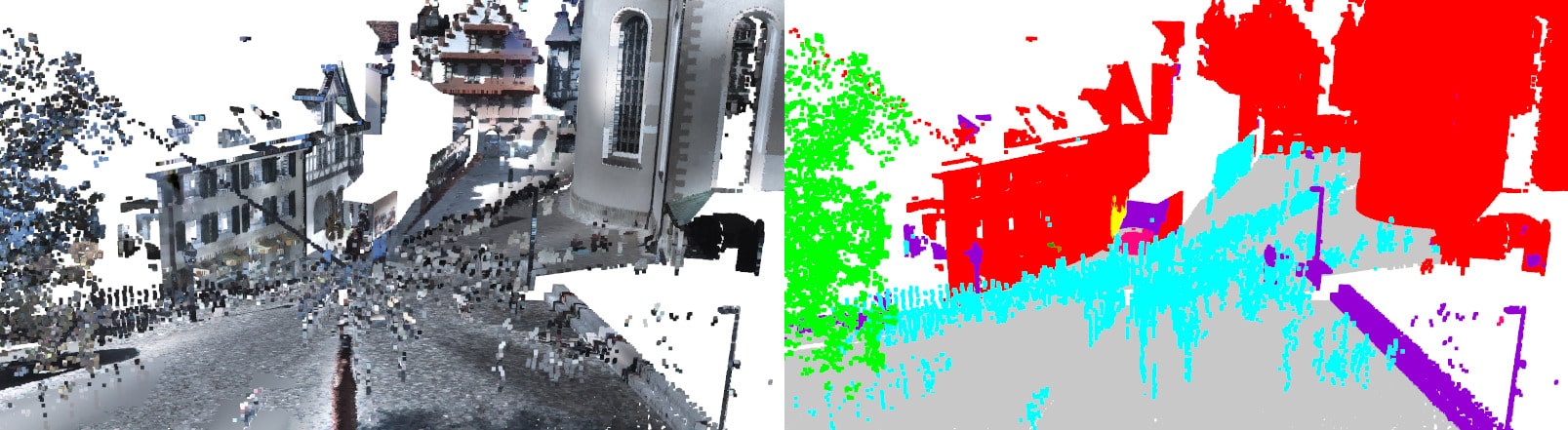}
\vspace{8pt}
\includegraphics[trim={0 1.5cm 0 0},clip,width=\textwidth]{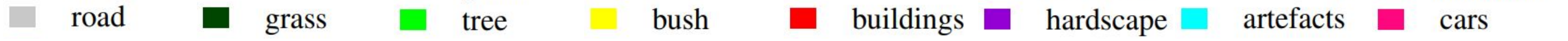}
\caption{Qualitative results of \nickname{} on the \textit{reduced-8} split of Semantic3D. From left to right: full RGB colored point clouds, predicted semantic labels of full point clouds, detailed view of colored point clouds, detailed view of predicted semantic labels. Note that the ground truth of the test set is not publicly available.}
\label{fig:reduced8-additional}
\end{figure*}

More qualitative results of \nickname{} on the Semantic3D \cite{Semantic3D} dataset (\textit{reduced-8}) are shown in Figure  \ref{fig:reduced8-additional}.

\section{Additional Results on SemanticKITTI}
\begin{figure*}[t]
\centering
\includegraphics[width=1\textwidth]{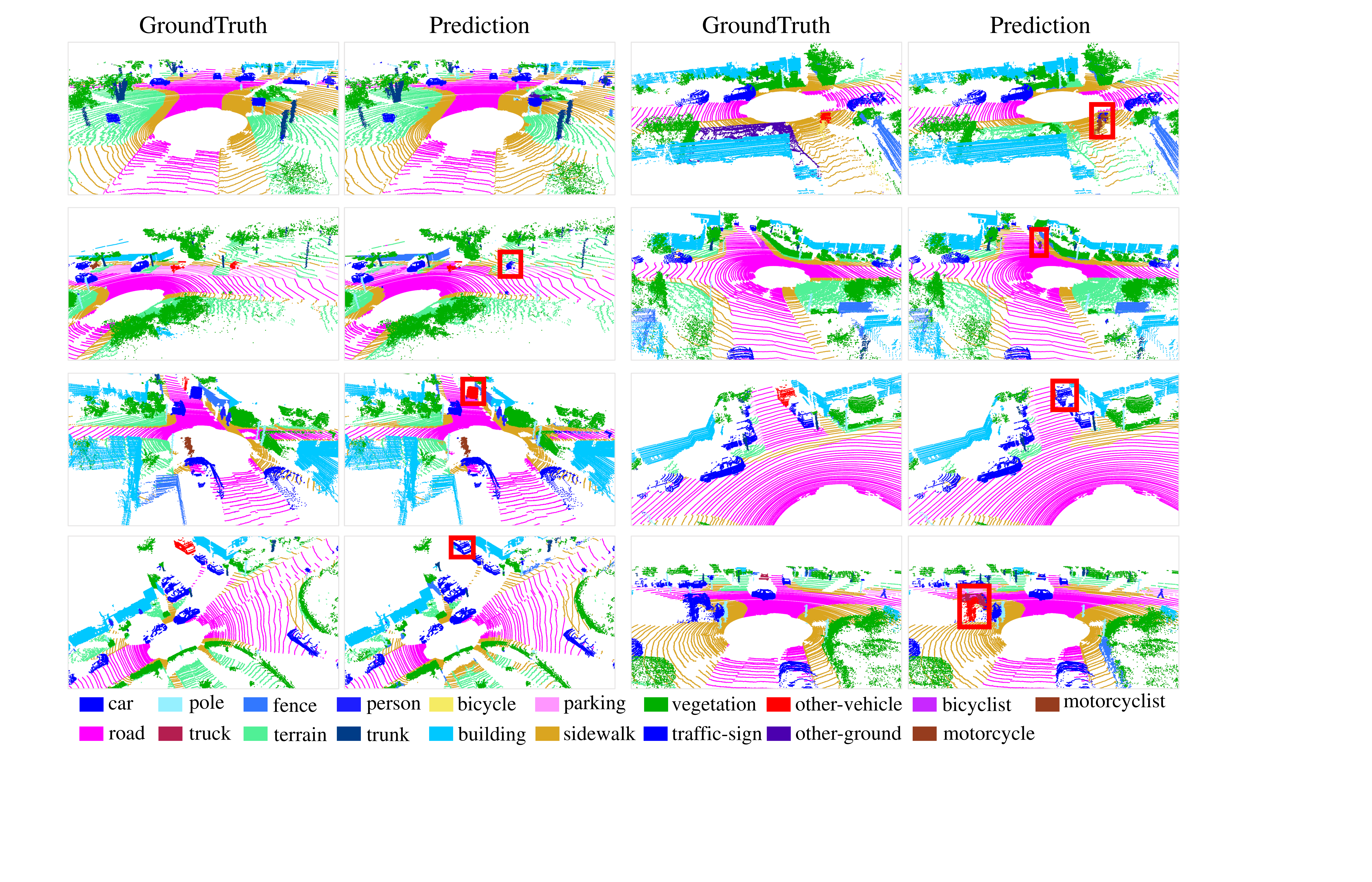}
\caption{Qualitative results of \nickname{} on the validation split of SemanticKITTI \cite{behley2019semantickitti}. Red boxes show the failure cases.}
\label{KITTI_supplementary}
\end{figure*}

Figure \ref{KITTI_supplementary} shows more qualitative results of our \nickname{} on the validation set of SemanticKITTI \cite{behley2019semantickitti}. The red boxes showcase the failure cases. It can be seen that, the points belonging to \textit{other-vehicle} are likely to be misclassified as \textit{car}, mainly because the partial point clouds without colors are extremely difficult to be distinguished between the two similar classes. In addition, our approach tends to fail in several minority classes such as \textit{bicycle}, \textit{motorcycle}, \textit{bicyclist} and \textit{motorcyclist}, due to the extremely imbalanced point distribution in the dataset. For example, the number of points for \textit{vegetation} is 7000 times more than that of \textit{motorcyclist}.

\section{Additional Results on S3DIS}
\label{Add_S3DIS}
\begin{figure*}[t]
\centering
\includegraphics[width=0.7\textwidth]{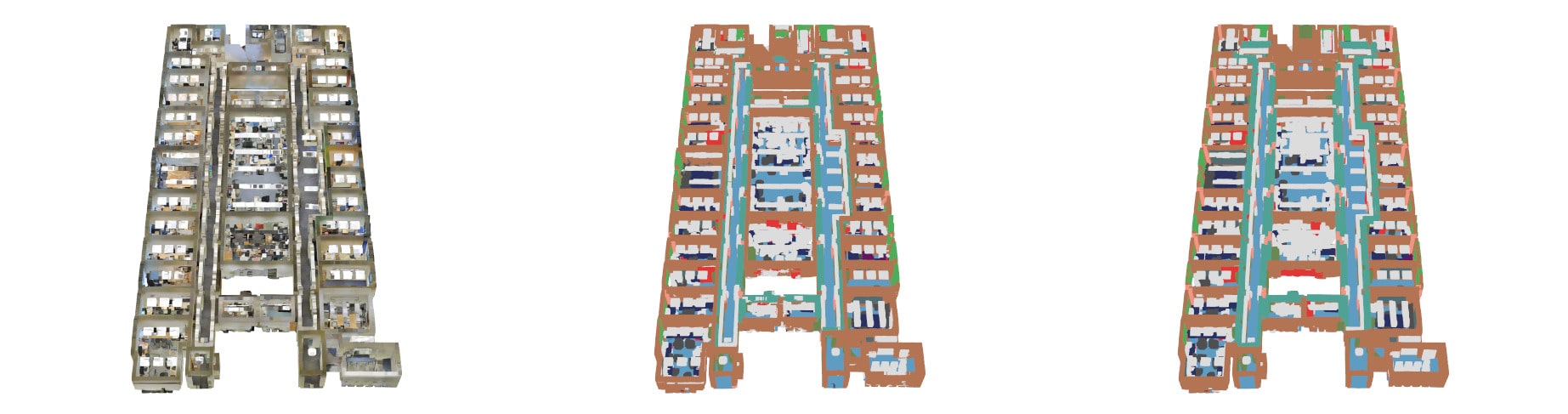}
\includegraphics[width=0.7\textwidth]{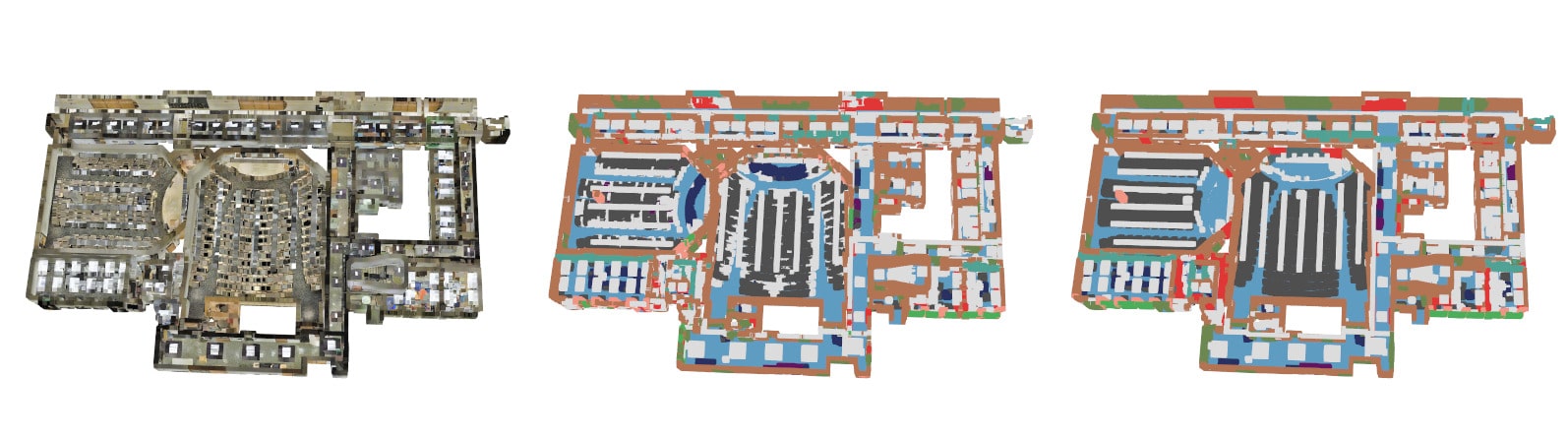}
\includegraphics[width=0.7\textwidth]{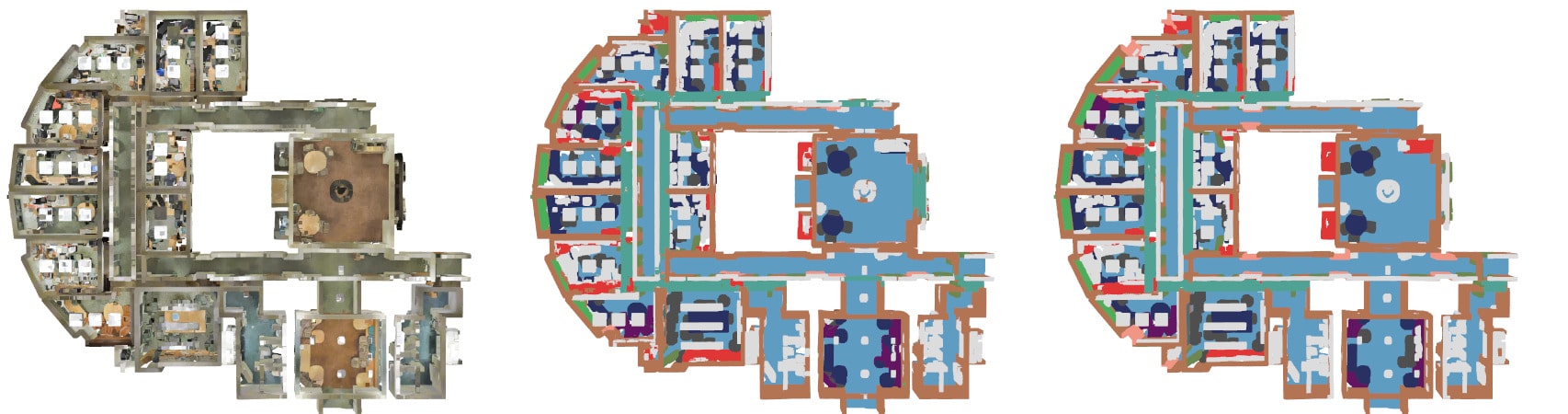}
\includegraphics[width=0.7\textwidth]{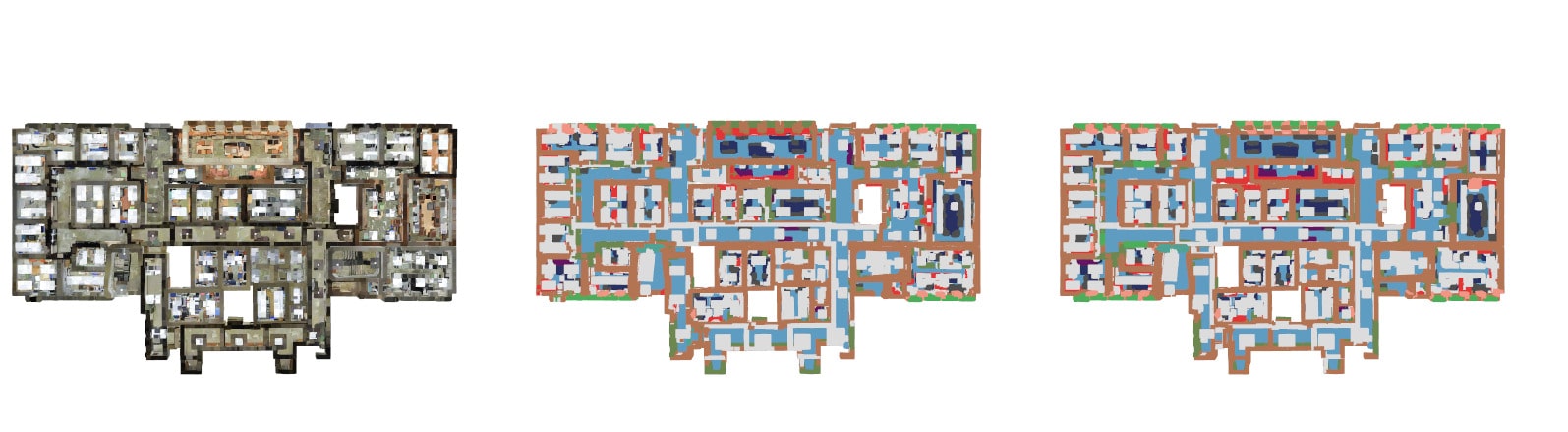}
\includegraphics[width=0.7\textwidth]{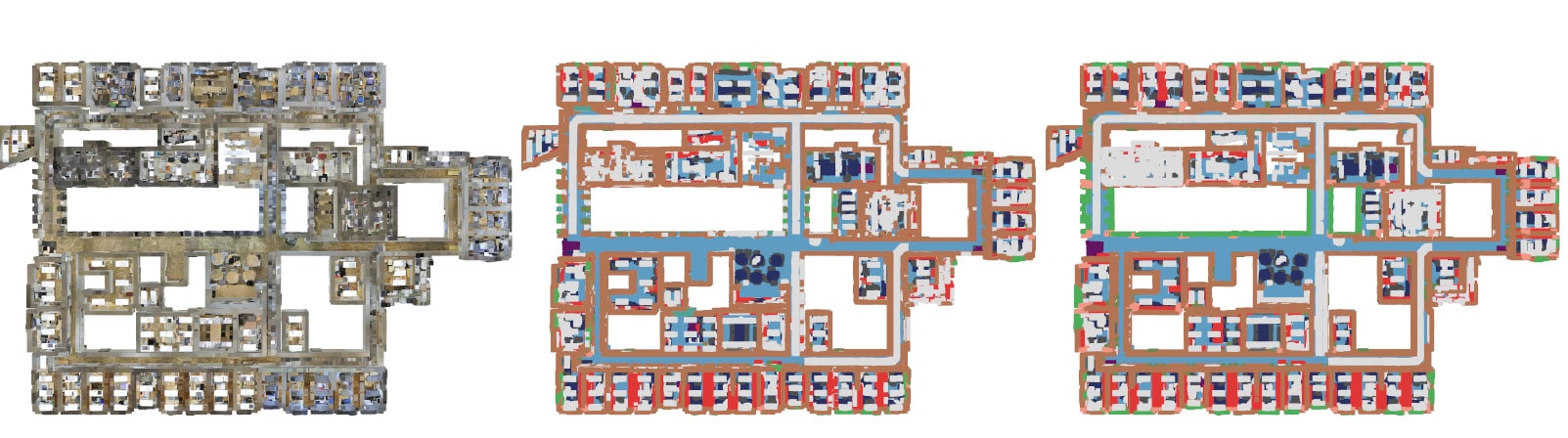}
\includegraphics[width=0.7\textwidth]{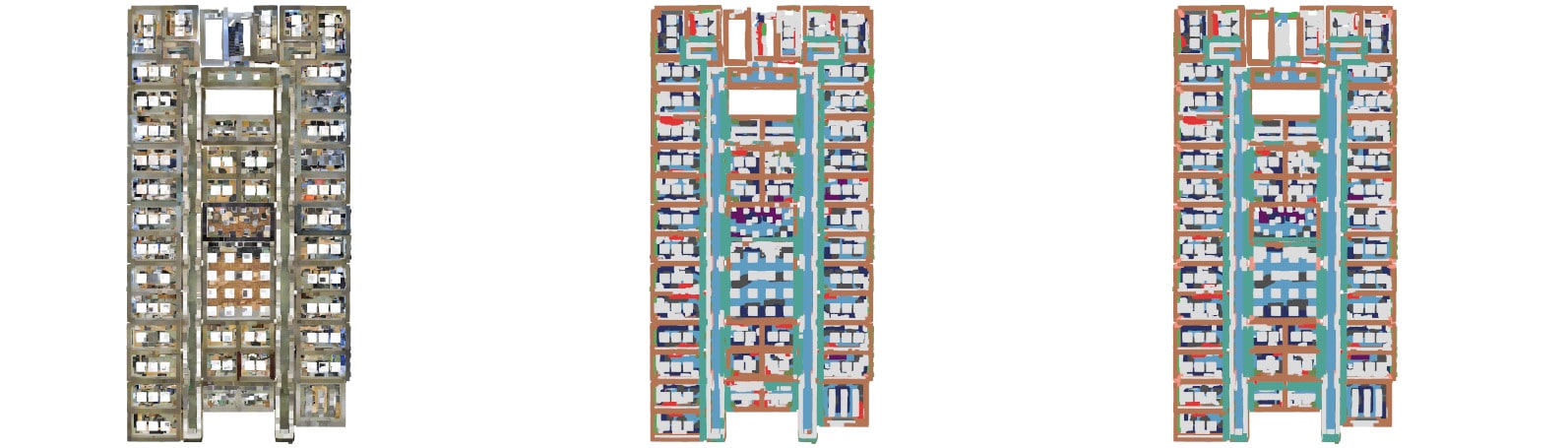}
\includegraphics[width=\textwidth]{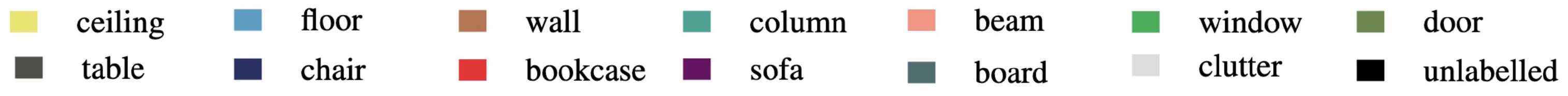}
\caption{Semantic segmentation results of our \nickname{}  on the complete point clouds of Areas 1-6 in S3DIS. Left: full RGB input cloud; middle: predicted labels; right: ground truth.}
\label{fig:s3dis-additional}
\end{figure*}

We report the detailed 6-fold cross validation results of our \nickname{} on S3DIS \cite{2D-3D-S} in Table \ref{tab:S3DIS}. Figure \ref{fig:s3dis-additional} shows more qualitative results of our approach.

 \begin{table*}[th]
\centering
\caption{Quantitative results of different approaches on S3DIS \cite{2D-3D-S} (6-fold cross-validation). Overall Accuracy (OA, \%), mean class Accuracy (mAcc, \%), mean IoU (mIoU, \%), and per-class IoU (\%) are reported.}
\label{tab:S3DIS}
\resizebox{\textwidth}{!}{%
\begin{tabular}{rcccccccccccccccc}
\toprule[1.0pt]
& OA(\%) & mAcc(\%)& mIoU(\%) & ceil. & floor & wall & beam & col. & wind. & door & table & chair & sofa & book. & board & clut. \\ 
\toprule[1.0pt]
PointNet \cite{qi2017pointnet} & 78.6 & 66.2 & 47.6 & 88.0 & 88.7 & 69.3 & 42.4 & 23.1 & 47.5 & 51.6 & 54.1 & 42.0 & 9.6 & 38.2 & 29.4 & 35.2 \\
RSNet \cite{RSNet} & - & 66.5 & 56.5 & 92.5 & 92.8 & 78.6 & 32.8 & 34.4 & 51.6 & 68.1 & 59.7 & 60.1 & 16.4 & 50.2 & 44.9 & 52.0 \\
3P-RNN \cite{3PRNN} & 86.9 & - & 56.3 & 92.9 & 93.8 & 73.1 & 42.5 & 25.9 & 47.6 & 59.2 & 60.4 & 66.7 & 24.8 & 57.0 & 36.7 & 51.6 \\
SPG \cite{landrieu2018large}& 86.4 & 73.0 & 62.1 & 89.9 & 95.1 & 76.4 & 62.8 & 47.1 & 55.3 & 68.4 & \textbf{73.5} & 69.2 & 63.2 & 45.9 & 8.7 & 52.9 \\
PointCNN \cite{li2018pointcnn} & \textbf{88.1} & 75.6 & 65.4 & \textbf{94.8} & \textbf{97.3} & 75.8 & 63.3 & 51.7 & 58.4 & 57.2 & 71.6 & 69.1 & 39.1 & 61.2 & 52.2 & 58.6 \\ 
PointWeb \cite{pointweb} & 87.3 & 76.2 & 66.7 & 93.5 & 94.2 & 80.8 & 52.4 & 41.3 & 64.9 & 68.1 & 71.4 & 67.1 & 50.3 & 62.7 & 62.2 & 58.5  \\
ShellNet \cite{zhang2019shellnet} & 87.1  & - & 66.8 & 90.2 & 93.6 & 79.9 & 60.4 & 44.1 & 64.9 & 52.9 & 71.6 & \textbf{84.7} & 53.8 & 64.6 & 48.6 & 59.4 \\
KPConv \cite{thomas2019kpconv} &- & 79.1 &\textbf{70.6} &93.6 &92.4 & \textbf{83.1} & \textbf{63.9} & \textbf{54.3} & \textbf{66.1} & \textbf{76.6} &57.8 &64.0 & \textbf{69.3} & \textbf{74.9} &61.3 & \textbf{60.3} \\
\textbf{\nickname{} (Ours)} & 88.0 & \textbf{82.0} & 70.0 & 93.1 & 96.1 & 80.6 & 62.4 & 48.0 & 64.4 & 69.4 & 69.4 & 76.4 & 60.0 & 64.2 & \textbf{65.9} & 60.1 \\
\bottomrule[1.0pt]\end{tabular}%
}
\end{table*}

\section{Video Illustration}
We provide a video to show qualitative results of our \nickname{} on both indoor and outdoor datasets, which can be viewed at \url{https://www.youtube.com/watch?v=Ar3eY_lwzMk&t=9s}.

\end{appendices}